\documentclass[10pt,twocolumn,letterpaper]{article}

\usepackage{cvpr}              

\cvprfinalcopy 

\usepackage{graphicx}
\usepackage{amsmath}
\usepackage{amssymb}
\usepackage{booktabs}

\usepackage{url}            
\usepackage{booktabs}       
\usepackage{amsfonts}       
\usepackage{nicefrac}       
\usepackage{microtype}      
\usepackage{multirow}
\usepackage{easyReview} 

\usepackage{latexsym}
\usepackage{algorithm}
\usepackage{algorithmic}

\usepackage{subcaption}

\usepackage{xcolor}

\usepackage[pagebackref,breaklinks,colorlinks]{hyperref}
\usepackage[capitalize]{cleveref}
\crefname{section}{Sec.}{Secs.}
\Crefname{section}{Section}{Sections}
\Crefname{table}{Table}{Tables}
\crefname{table}{Tab.}{Tabs.}

\def\cvprPaperID{1979} 

\begin{document}
\title{Towards Automated Polyp Segmentation Using Weakly- and Semi-Supervised Learning and Deformable Transformers}

\author{
Guangyu Ren{*}\ \ \ \ Michalis Lazarou\thefootnote{*}\ \ \ \ Jing Yuan\thefootnote{*} \ \ \ \ Tania Stathaki\\
Imperial College London\\
}

\maketitle
\begin{abstract}
Polyp segmentation is a crucial step towards computer-aided diagnosis of colorectal cancer. However, most of the polyp segmentation methods require pixel-wise annotated datasets. Annotated datasets are tedious and time-consuming to produce, especially for physicians who must dedicate their time to their patients.  We tackle this issue by proposing a novel framework that can be trained using only weakly annotated images along with exploiting unlabeled images. To this end, we propose three ideas to address this problem, more specifically our contributions are: 1) a novel sparse foreground loss that suppresses false positives and improves weakly-supervised training, 2) a batch-wise weighted consistency loss utilizing predicted segmentation maps from identical networks trained using different initialization during semi-supervised training, 3) a deformable transformer encoder neck for feature enhancement by fusing information across levels and flexible spatial locations.



Extensive experimental results demonstrate the merits of our ideas on five challenging datasets
outperforming some state-of-the-art fully supervised models. Also, our framework can be utilized to fine-tune models trained on natural image segmentation datasets drastically improving their performance for polyp segmentation and impressively demonstrating superior performance to fully supervised fine-tuning.
\def\thefootnote{*}\footnotetext{These authors contributed equally to this work.}

\end{abstract}

\section{Introduction}
\label{sec:intro}

Automated medical image segmentation has attracted interest in recent years due to its potential to significantly reduce the workload of physicians by being used as a supporting tool for a physician's diagnosis. Due to the rapid development of deep learning \cite{DL_book}, the current state-of-the-art image segmentation methods utilize deep learning techniques and medical image segmentation has been no exception. 

However, one of the bottlenecks of deep learning techniques is their reliance on large, well-annotated datasets. Annotating datasets for image segmentation is particularly time-consuming since pixel-wise annotations must be provided which requires significant manual labour. While in standard image segmentation anyone can annotate a dataset, in medical images, annotations must be provided by expert physicians that are trained to detect lesions in these images. This is a significant limitation for automated medical image segmentation since physicians do not have time to dedicate to annotating images.



To address this issue and save physicians' valuable time we propose a novel framework for medical image segmentation. Our framework can be trained using only weakly annotated images and unlabeled images. These weak annotations include only information regarding where the foreground and background pixels are located. 


\begin{figure}
\minipage{0.1\textwidth}
  \centering
  \includegraphics[width=\linewidth]{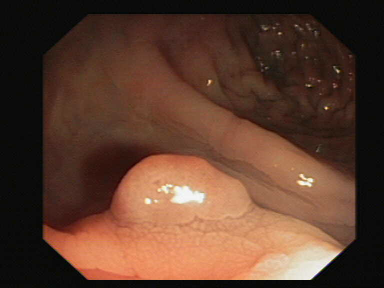}
    (a)
\endminipage\hfill
\minipage{0.1\textwidth}
  \centering
  \includegraphics[width=\linewidth]{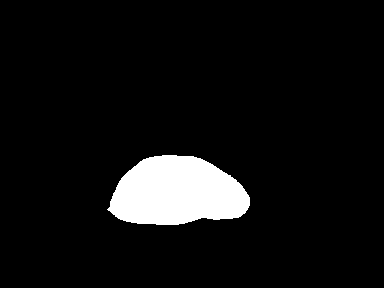}
    (b) 
\endminipage\hfill
\minipage{0.1\textwidth}
  \centering
  \includegraphics[width=\linewidth]{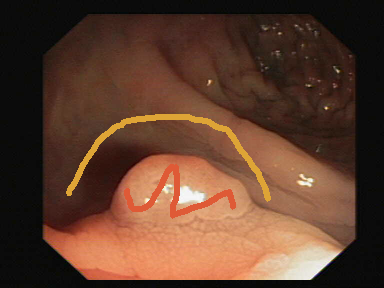}
    (c)
\endminipage\hfill
\minipage{0.1\textwidth}
  \centering
  \includegraphics[width=\linewidth]{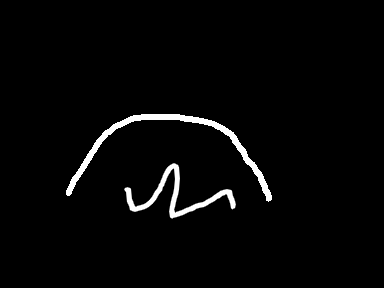}
    (d)
\endminipage\hfill

\minipage{0.1\textwidth}
  \centering
  \includegraphics[width=\linewidth]{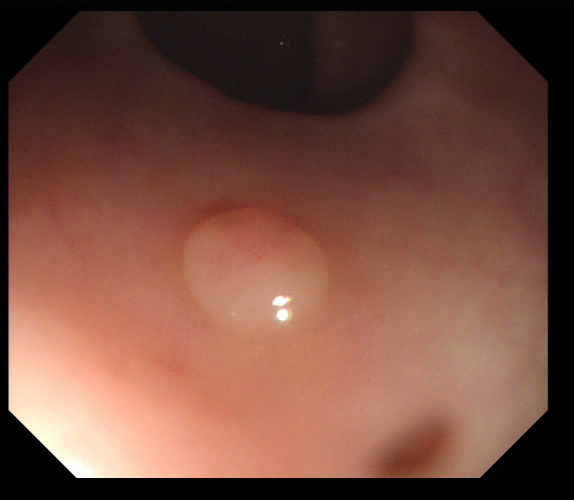}
    (e)
\endminipage\hfill
\minipage{0.1\textwidth}
  \centering
  \includegraphics[width=\linewidth]{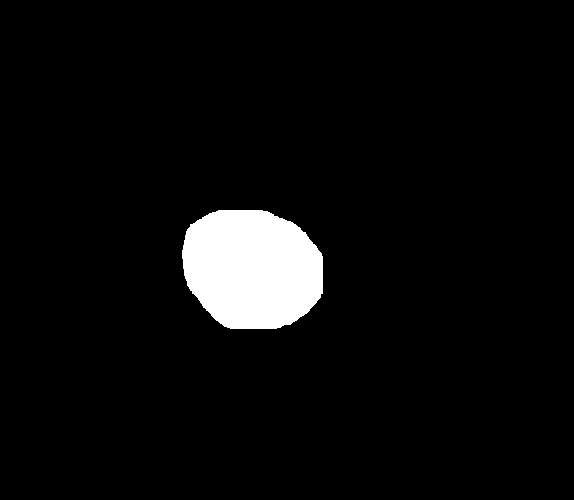}
    (f)
\endminipage\hfill
\minipage{0.1\textwidth}
  \centering
  \includegraphics[width=\linewidth]{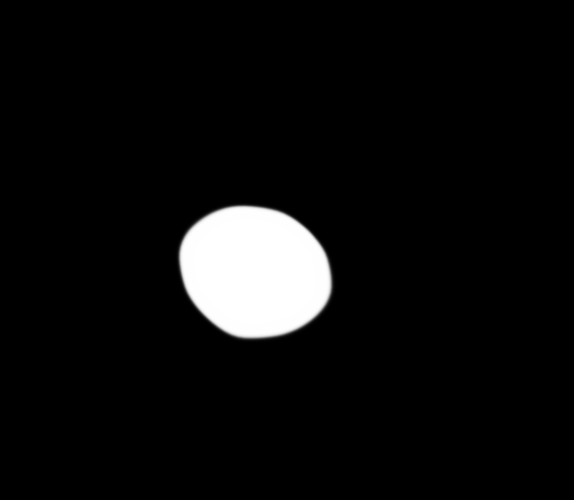}
    (g)
\endminipage\hfill
\minipage{0.1\textwidth}
  \centering
  \includegraphics[width=\linewidth]{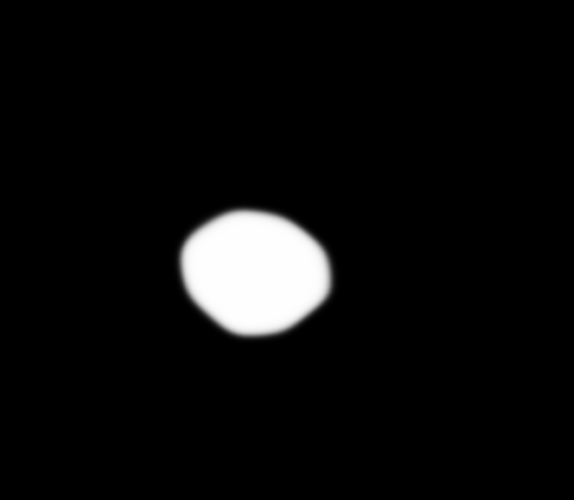}
    (h)
\endminipage\hfill

\caption{Visualization of weak annotations. (a) RGB image of the training data. (b) Original ground truth. (c) Foreground and background. (d) Our weak annotations. (e) RGB image of the testing data. (f) Corresponding ground truth. (g) Predicted map of fully supervised training manner. (h) Predicted map of our methods.}
\label{fig:l new label examples}
\end{figure}

Specifically, we leverage our framework on polyp segmentation, which aims at detecting and segmenting polyps for the early diagnosis of colorectal cancer. Current research \cite{fan2020pranet, wu2021collaborative, fang2019sfa} still relies on complete polyp annotations to achieve accurate detection performance. Under this circumstance, we relabel the training dataset with weak annotations by simply drawing sketches.
Only around 1.9$\%$ of the total pixels of all images in the whole dataset are labeled. The annotations simply need to indicate the foreground (polyp region) and the background (non-polyp region), making this annotation strategy very efficient for physicians to use without sacrificing a lot of time. Our weak annotations can be seen in Figure \ref{fig:l new label examples}(c) and (d) annotating the polyp region and the non-polyp region with two simple lines in direct contrast to the original ground truth segmentation maps that require pixel-wise careful annotation.

Our proposed framework consists of a two-stage training regime. In the first stage, a model is trained using a weakly-supervised training paradigm while in the second stage we train the model using a semi-supervised learning paradigm. Also, as part of our framework, we propose a novel architectural component that is used for feature enhancement. 

\begin{figure}
\minipage{0.15\linewidth}
  \centering
  \includegraphics[width=\linewidth]{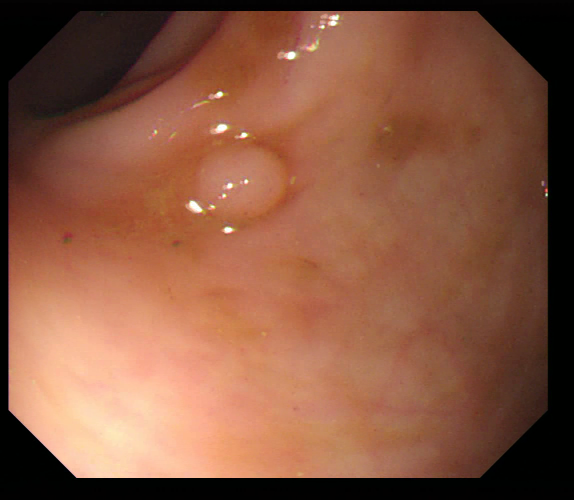}
\endminipage\hfill
\minipage{0.15\linewidth}
  \centering
  \includegraphics[width=\linewidth]{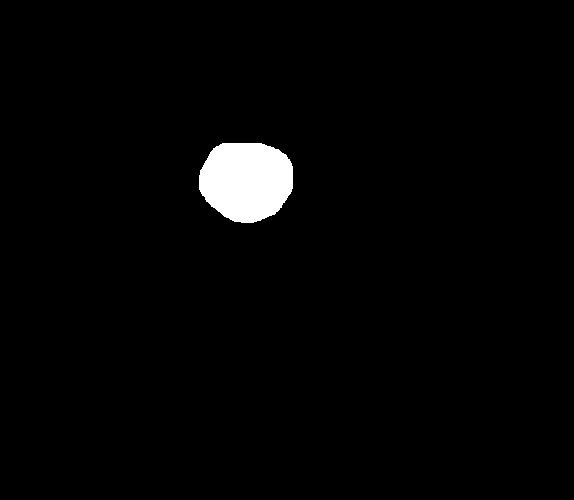}
\endminipage\hfill
\minipage{0.15\linewidth}
  \centering
  \includegraphics[width=\linewidth]{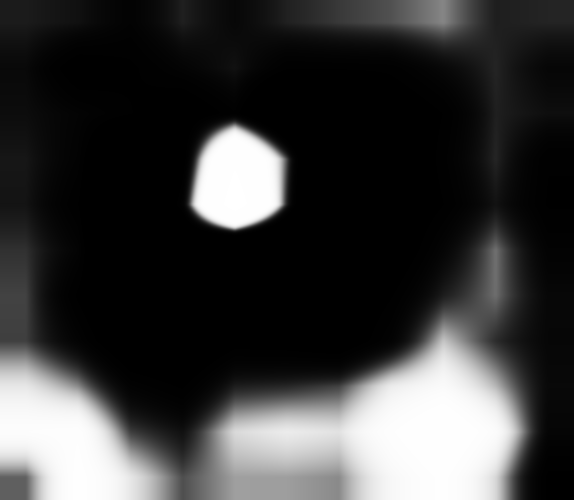} 
\endminipage\hfill
\minipage{0.15\linewidth}
  \centering
  \includegraphics[width=\linewidth]{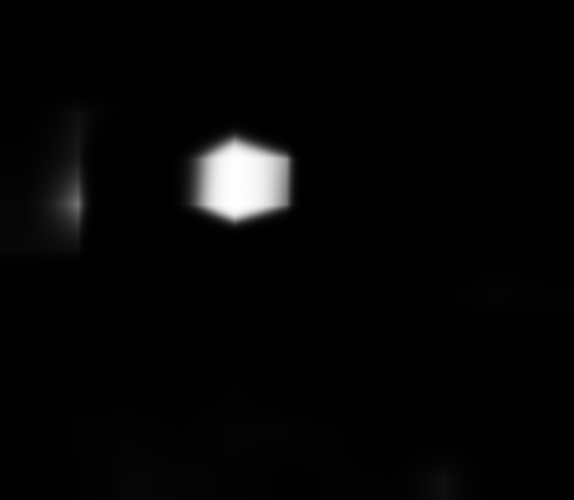} 
\endminipage\hfill
\minipage{0.15\linewidth}
  \centering
  \includegraphics[width=\linewidth]{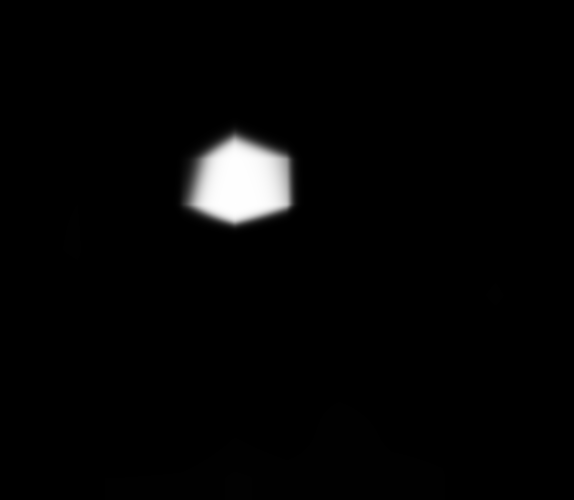}
\endminipage\hfill
\minipage{0.15\linewidth}
  \centering
  \includegraphics[width=\linewidth]{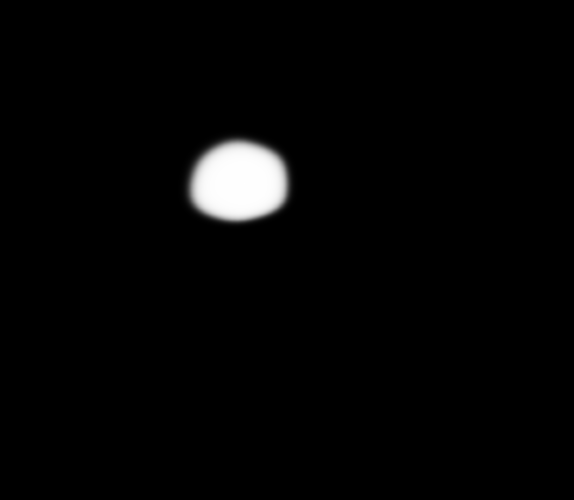}
\endminipage\hfill

\minipage{0.15\linewidth}
  \centering
  \includegraphics[width=\linewidth]{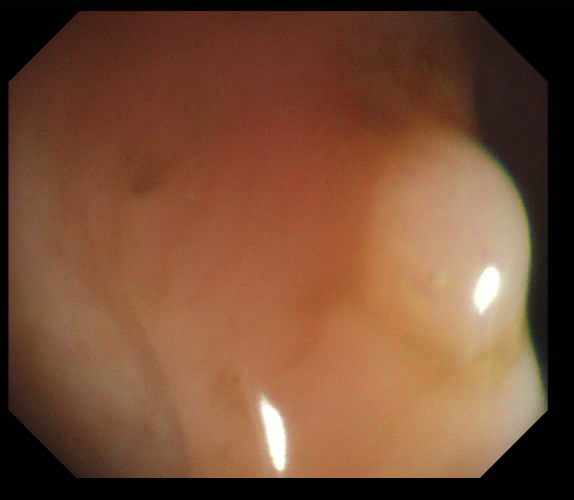}
   (a)
\endminipage\hfill
\minipage{0.15\linewidth}
  \centering
  \includegraphics[width=\linewidth]{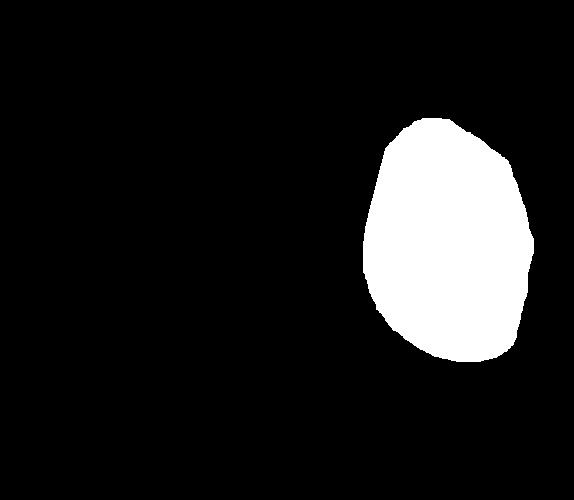}
   (b)
\endminipage\hfill
\minipage{0.15\linewidth}
  \centering
  \includegraphics[width=\linewidth]{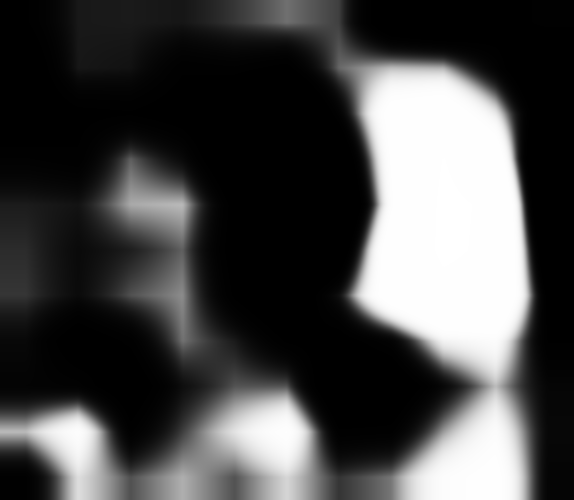} 
   (c)
\endminipage\hfill
\minipage{0.15\linewidth}
  \centering
  \includegraphics[width=\linewidth]{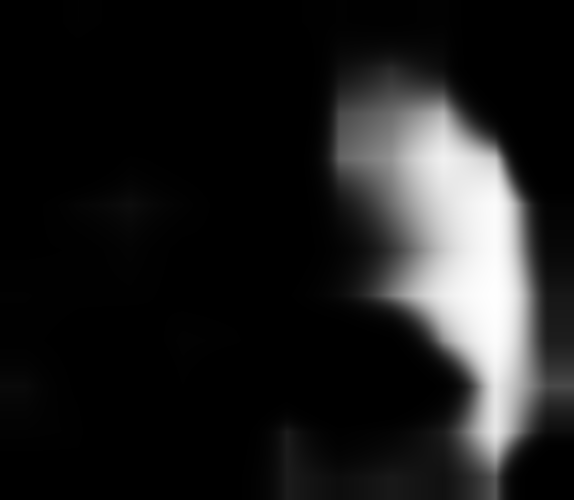} 
   (d)
\endminipage\hfill
\minipage{0.15\linewidth}
  \centering
  \includegraphics[width=\linewidth]{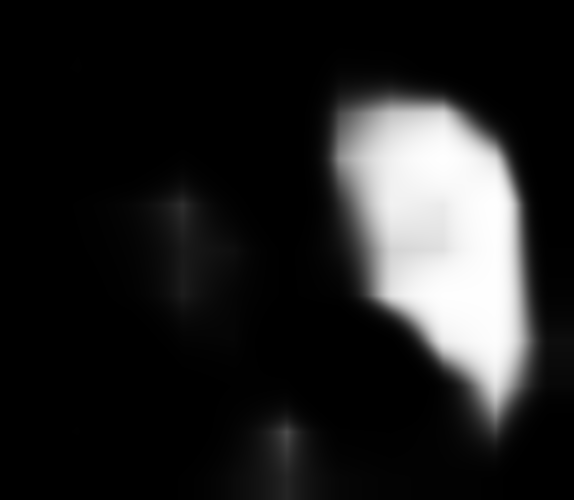}
   (e)
\endminipage\hfill
\minipage{0.15\linewidth}
  \centering
  \includegraphics[width=\linewidth]{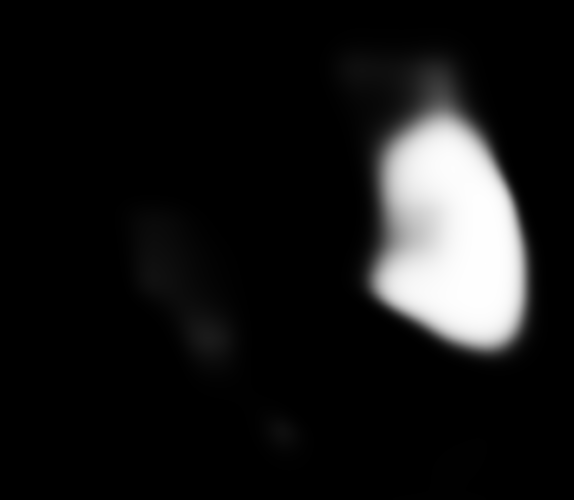}
   (f)
\endminipage\hfill

\caption{Visual comparison of ablation study. (a) RGB image. (b) Original ground truth (c) backbone. (d) +sparse foreground loss. (e) +semi with weighted consistency loss. (f) +DTEN.}
\label{fig:false_positives}
\end{figure}

During the weakly-supervised training stage, we propose a novel weakly-supervised loss function that addresses a key limitation of weakly-supervised training techniques, that of numerous false positives \cite{zhang2020weakly,yu2021structure}. Since weak annotations contain only a fraction of the polyp region, training models by partial cross-entropy loss \cite{tang2018normalized} could cause a large number of false positives as shown in Figure \ref{fig:false_positives}(c). 
A previous work \cite{zhang2020weakly} attempted to address the problem of false positives using an auxiliary edge detection network supervising the model to align image edges with the predicted segmentation map boundaries. However, this method complicates the training process and relies on auxiliary networks.
To address this problem in a simple way, we propose a novel sparse foreground loss function that suppresses false positives and refines the rough predicted segmentation maps (Figure \ref{fig:false_positives}(d)). 

In addition, because of the weakly-supervised training, inconsistent segmentation maps can be generated by two identical models trained the same way (Figure \ref{fig: weak issues}(b) and (c)). To exploit the prior knowledge of the predicted map, we propose a batch-wise weighted consistency loss to utilize two predicted segmentation maps during semi-supervised training.

Lastly in order to improve the accuracy performance even further, we propose a Deformable Transformer Encoder Neck (DTEN) which leverages a multi-scale deformable self-attention encoder along with a novel progressive compensation sequence for feature enhancement. The merit of each of our ideas can be visualized from Figure \ref{fig:false_positives} (c)-(f), each idea improves consistently the performance. We name our novel framework Weakly- and Semi-supervised Deformable Segmentation network, in short, \textbf{WS-DefSegNet}.

To summarize, the contributions of our work are the following:
\begin{itemize}
    \item We are the first, to the best of our knowledge, to propose a weakly- and semi-supervised training framework for efficient polyp segmentation. To this end, we propose a novel sparse foreground loss and a batch-wise weighted consistency loss.
    \item We propose DTEN, a novel progressive multi-scale architecture with a self-attention mechanism for feature enhancement that significantly improves the performance of WS-DefSegNet.
    \item We created the first, to the best of our knowledge, weakly annotated polyp segmentation dataset \textbf{W-Polyp}.  We are planning on making it publicly available as a way to promote research in this direction. 
   
    \item We provide extensive experimental results showing the merits of WS-DefSegNet. Also, we show the transferability of our framework by adapting models that were trained on completely different datasets and different tasks.

\end{itemize}

\section{Related Work}
\label{sec:related}

\paragraph{Medical Image Segmentation}
\label{sec:sod}
Medical image segmentation aims at identifying lesion areas which indicate potential diseases in the human tissue. Deep learning methods have achieved compelling performance due to a fully supervised training paradigm.  U-net \cite{ronneberger2015unet} designs a U-shape architecture built on fully convolutional networks to capture context features and gradually segment biomedical images with precise localization. Analogously, CE-net \cite{gu2019cenet} proposes an encoder-decoder structure with a dense atrous convolution block for medical segmentation and \cite{li2020memory} inherits the U-net framework and proposes a non-local context-guided mechanism to capture long-range pixel-wise dependencies in features for tumor segmentation. 

More specifically for polyp segmentation, Pranet \cite{fan2020pranet} proposes a recurrent reverse attention module to mine boundary cues and a parallel partial decoder. Other approaches \cite{zhou2018unet++, jha2019resunet++, fang2019sfa} have also been proposed with the overwhelming majority focusing on fully supervised training. In contrast, our framework only uses weak annotations and outperforms some of the aforementioned methods.



\paragraph{Weakly-supervised Segmentation}

\label{sec:polyp}
To avoid tediously labeling pixel-wise annotations, image segmentation is encouraged through the use of inexpensive labels, formulating the weakly-supervised training paradigm using image-level labels and weak labels. Ahn \cite{ahn2019weakly} proposes an IRNet to estimate rough areas of individual instances and detect boundaries with image-level class labels. Chen \cite{chen2020weakly} explicitly explores object boundaries through coarse localization and proposes a BENet to further excavate more
object boundaries.
Zhang \cite{zhang2020weakly} leverages scribble annotations by relabeling an existing salient object detection dataset and further adopting an auxiliary edge detection task to explicitly provide edge supervision on the final output. Yu \cite{yu2021structure} designs a local coherence loss to improve boundary localization and a structure consistency loss to further enhance the model's generalization ability. However, the aforementioned methods use auxiliary networks and focus on excavating edge information, while in our work we propose an effective weakly-supervised loss function for polyp segmentation.



\paragraph{Semi-supervised Learning}
\label{sec:semi_supervised}

Semi-supervised learning addresses the research question of exploiting unlabeled data together with labeled data to improve the performance of a model. A line of research attracting attention in recent years is that of consistency regularization, where the main idea is to enforce similar predictions between two cases, either two different augmentations of the same image or the same image but predictions made from two different networks \cite{ss_temporal, ss_transformations, mean_teacher}. 
Pseudo-labeling unlabeled data and using them in the training process is another promising direction, for example, \cite{pseudo_ss} uses the current network to assign pseudo-labels to the unlabeled data while \cite{deep_lp} uses label propagation to exploit the underlying manifold structure of the data to assign pseudo-labels. Other influential works such as MixMatch \cite{mixmatch} and ReMixMatch \cite{remixmatch} incorporate many ideas together, such as using data augmentation consistency, applying mixUp regularization \cite{mixup} and distribution alignment \cite{distribution_alignment}. For further information regarding semi-supervised learning, we refer the reader to \cite{semisupervised_learning}.


Semi-supervised learning is utilized successfully for segmentation tasks.
The authors of \cite{ss_segmentation_augmentation} exploit strong augmentations effectively by designing a distribution-specific batch normalization since previous attempts failed due to the large distribution disparity caused by strong augmentations. Other data augmentation based methods \cite{cut_paste, instaboost, classmix} focus on cutting and pasting annotated objects from images to new backgrounds. 

Focusing on polyp segmentation, Wu \cite{wu2021collaborative} employs two collaborative segmentation networks for semi-supervised polyp segmentation and two discriminators to minimize the impact of the imbalance problem between labeled and unlabeled data. However, in contrast to our work they use a fully annotated subset of polyps while we only use weak annotations.



\paragraph{Vision Transformers in Medical Image Segmentation}
\label{sec:transformers}
Vision transformers have been extensively applied to medical image segmentation owing to their capability to incorporate global features while maintaining high resolution. 
They can be used to establish effective backbones to improve lesion segmentation. \cite{ou2022patcher} stacks four Patcher blocks with vision transformer blocks as the core. \cite{lin2022contrans} encodes input image patches with multiple Swin transformer encoders \cite{liu2021swin} in parallel with the traditional CNN-based backbone.
Besides, transformers are used for feature fusion out of the backbone.
\cite{xing2022nestedformer} combines multi-modality features with the assistance of multiple transformer encoders and a single decoder for MRI brain image segmentation. \cite{wang2022DA-net} fuses the patch- and image-level features with three transformer encoders for retinal vessel segmentation. \cite{tang2022LeisionSegTransformer} appends six transformer encoder-decoders after the CNN backbone for lesion segmentation. 
To achieve efficient and accurate segmentation, we take the advantage of the multi-scale deformable transformer \cite{DETR} and only use a single transformer encoder to maximize inference speed.

\section{Efficient Polyp Segmentation}
\label{sec:method}

\subsection{W-Polyp Dataset}

\begin{figure}[h]
\centering
\includegraphics[width=0.8\linewidth]{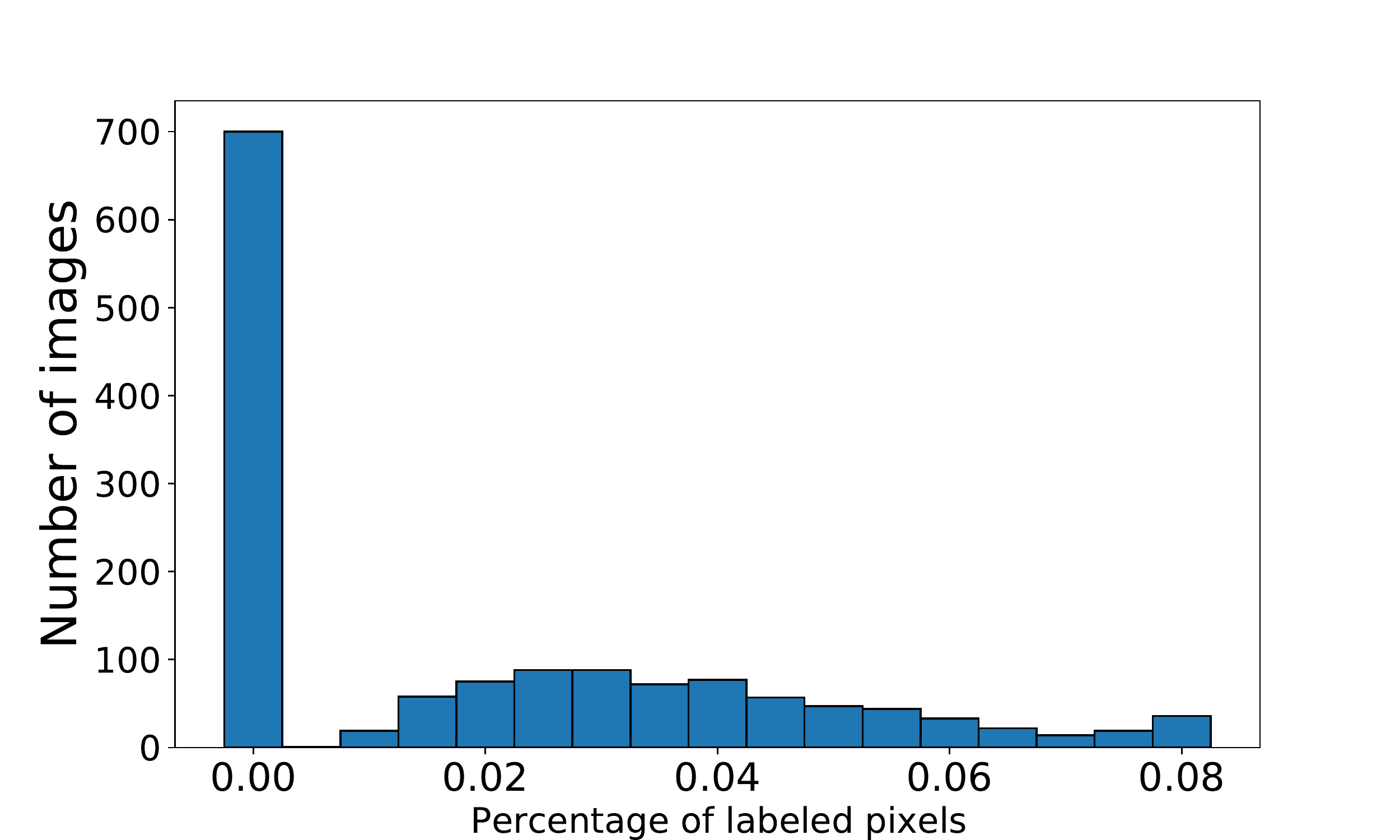}

\caption{ The histogram of the number of images versus the percentage of labeled pixels.}
\label{fig:label percent}
\end{figure}

As stated in section \ref{sec:intro},  we create the first weakly annotated dataset for polyp segmentation comprising of weakly annotated and unlabeled images named W-Polyp. W-Polyp is created by labeling the existing training data of \cite{fan2020pranet} which contains 1,450 images. We randomly selected and weakly annotated 750 images with simple sketches, including lines, scribbles and circles. Annotating an image in this way only takes 2 seconds. Additionally, unlike other weakly annotated datasets, the other 700 images are left unlabeled, maximizing labeling efficiency and enlarging the sparsity of the whole training data. Therefore, only around 1.9\% of pixels are labeled as foreground and background as shown in Figure \ref{fig:label percent}.

\begin{figure}[htbp]
\centering
\includegraphics[width=1\linewidth]{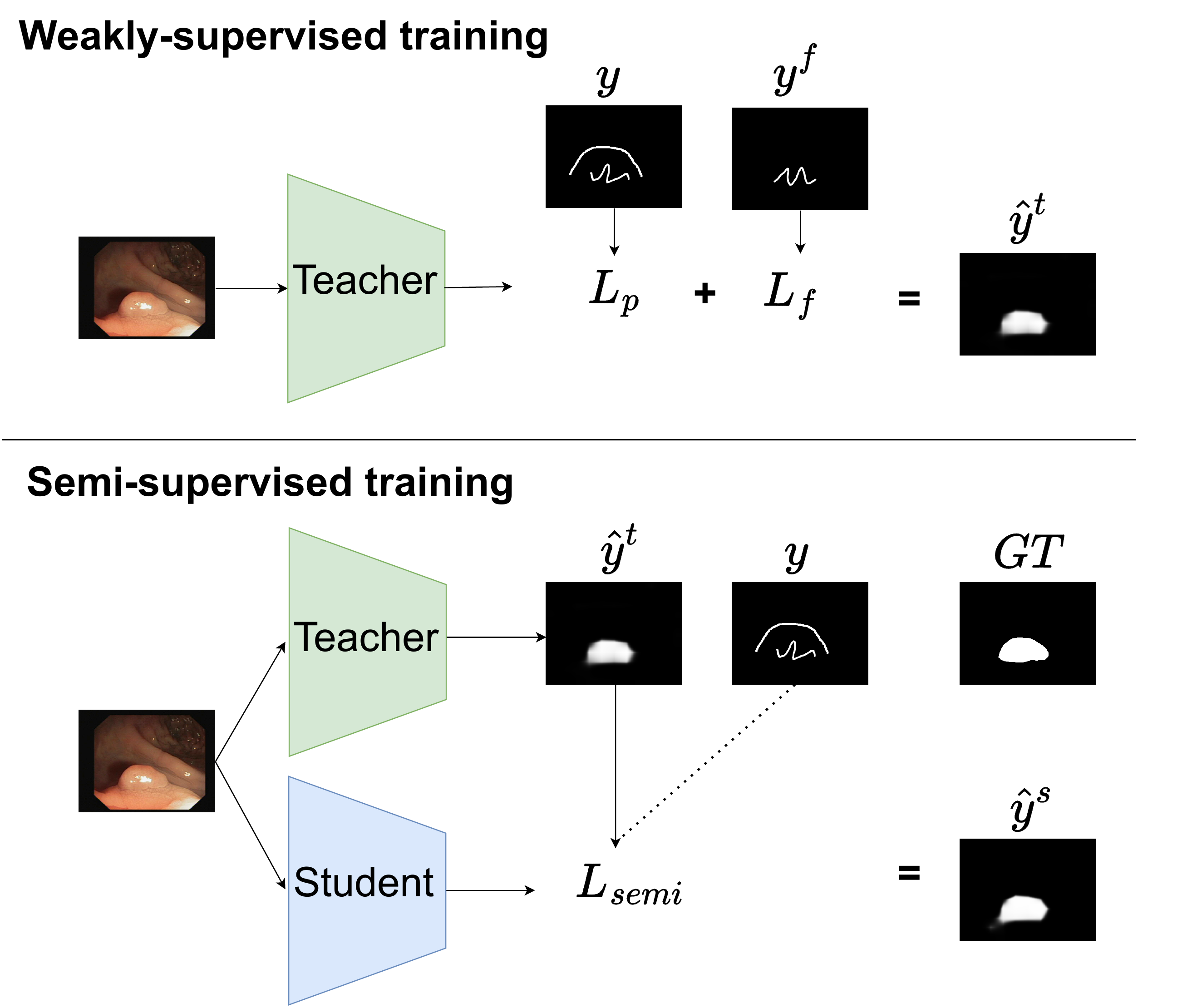}

\caption{The training procedure of our WS-DefSegNet. We first train the teacher network using a weakly-supervised paradigm as explained in \ref{sec:weakly_training}.  We train the final student network using a semi-supervised paradigm as explained in \ref{sec:semi_training}. $GT$ denotes the original ground truth map. Our WS-DefSegNet generates a satisfying segmentation map compared to the $GT$.}
\label{fig:framework_training}
\end{figure}

\begin{figure*}[htbp]
\centering
\includegraphics[width=.8\linewidth]{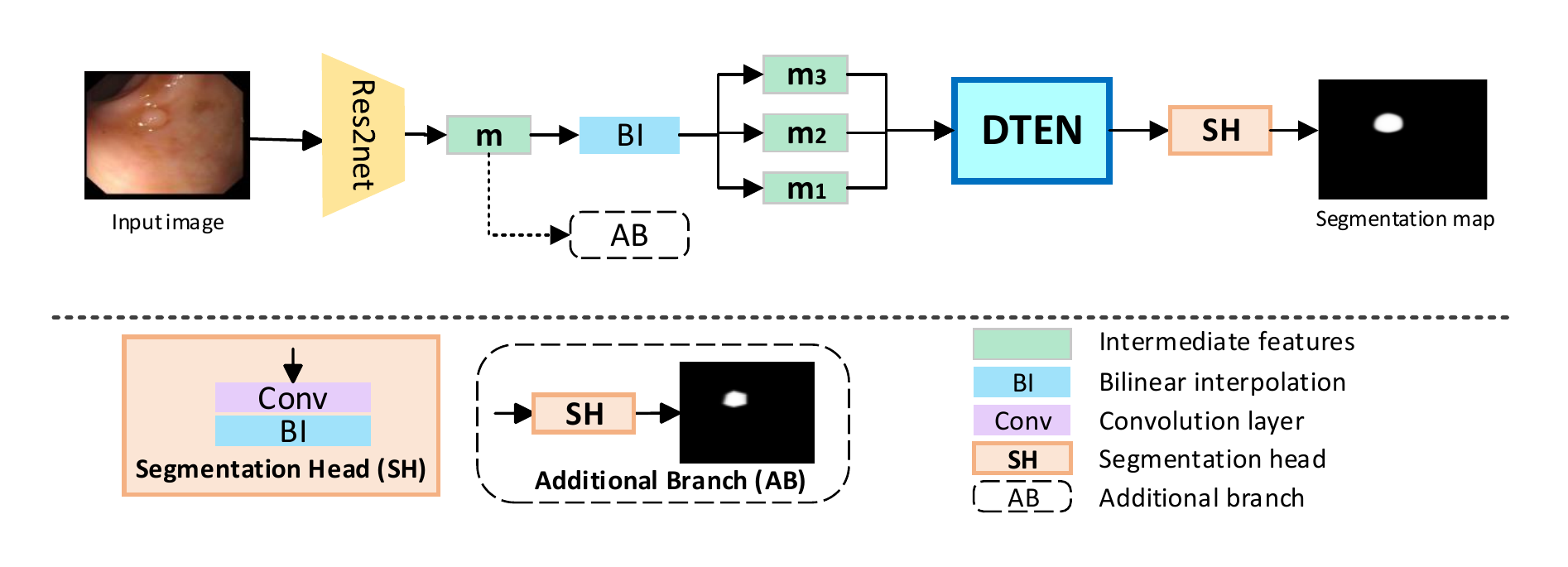}

\caption{The network architecture of our WS-DefSegNet. It utilizes the proposed Deformable Transformer Encoder Neck (DTEN) to enhance raw features produced by the last stage of the Res2net \cite{gao2019res2net}. Enhanced features are passed to a vanilla segmentation head. The additional branch in the dashed box only exists in the training stage.}
\label{fig:architechture}
\end{figure*}

\subsection{Method}

\subsubsection{Overview of WS-DefSegNet}

We propose the complete framework named WS-DefSegNet for efficient polyp segmentation. Our framework consists of two-training stages (Figure \ref{fig:framework_training}) and a network architecture (Figure \ref{fig:architechture}). The first training stage consists of a weakly-supervised training regime leveraging weakly annotated images while the second stage consists of a semi-supervised training regime leveraging both weakly annotated and unlabeled images. Regarding our network architecture, we propose a novel module, DETN, that uses deformable transformers for feature enhancement.  

\subsubsection{Problem Formulation}
We define the set of all images in our W-Polyp dataset as $X$. The subset of weakly-annotated images is defined as $X_l$ with their corresponding ground truth maps as $Y_l$ and the subset of unlabeled data as $X_u$, where $X_l \in X$, $X_u \in X$ and $X_l \cap X_u = \emptyset$. For every batch $B$, the labeled ground truth including foreground and background information is defined as $B_{l}$, while $B^f_{l}$ denotes only foreground annotations. 
  We denote our model $M_\theta$ where $\theta$ is the set of learnable parameters. $\hat{y}_i$ is the predicted segmentation map of the $i$-th image $x_i \in X$, $\hat{y}_i := M_{\theta}(x_i)$. 
  During the semi-supervised training stage, we use a teacher and a student model. We denote the teacher model as $M_\theta^t$ and the student model as $M_\theta^s$.

\subsubsection{Weakly-supervised Training}
\label{sec:weakly_training}

We define the partial cross-entropy loss utilized in \cite{zhang2020weakly} as follows:

\begin{equation}
 \centering
 L_{p}(\hat{y}_i, y_i) = \frac{1}{|B|}\sum_{y_i \in B_{l}} (y_{i}log \hat{y_{i}} + (1-y_{i})log (1- \hat{y_{i}}))
\end{equation}
where $y_{i}$ denotes the corresponding ground truth map with weak sketch annotations. Note that for all of the loss functions used in this work we average all per-pixel losses image-wise but omit this information from the equations to simplify our notation.

In order to mitigate the issue of false positives as described in section \ref{sec:intro}, we propose a novel loss function that utilizes only the foreground  pixels to supervise the model defined as:

\begin{equation}
 \centering
 L_{f}(\hat{y}_i, y_i^f)  = \frac{1}{|B|}\sum_{y^f_i \in B_{l}^f} (y_{i}^flog \hat{y_{i}} + (1-y^f_{i})log (1- \hat{y_{i}}))
\end{equation}
where $y^f_i$ indicates a ground truth map with only foreground annotations. Then the total loss for weakly-supervised learning can be defined as:

\begin{equation}
 \centering
 L_{weak}(\hat{y}_i, y_i, y_i^f) = L_{p}(\hat{y}_i, y_i) + \alpha \cdot L_{f}(\hat{y}_i, y_i^f)
 \label{eq:l_weak}
\end{equation}
where $\alpha$ is the weight of the sparse foreground loss. It is worth noting that $\alpha$ should be set appropriately. This is because small $\alpha$ makes the predicted segmentation map $\hat{y_{i}}$ contain many false positives, while large $\alpha$ forces the model to focus on the extremely sparse foreground pixels, leading to more false negatives. In this paper, it is set to 0.5, for further information please refer to the supplementary material.

\subsubsection{Semi-supervised Training}
\label{sec:semi_training}

We propose a teacher-student learning paradigm and train the teacher model as described in \ref{sec:weakly_training}. Using the teacher model, $M_{\theta}^t$ we assign pseudo-labels for every $x_i \in X$ defined as:

\begin{equation}
 \centering
 \hat{y}^{t}_i = M_{\theta}^t(x_i)
\end{equation}

In order to utilize the prior knowledge of the teacher model, $M_{\theta}^t$, for training the student model, $M_{\theta}^s$, we propose a batch-wise weighted consistency loss for semi-supervised learning:

\begin{equation}
 \centering
 L_{c}(\hat{y}^{s}_{i}, \hat{y}^{t}_i) = \frac{1}{|B|}  \sum_{i \in B} |\hat{y}^{s}_{i} - \hat{y}^{t}_i|
\end{equation}
where $\hat{y}^{s}_{i}$ refers to the predicted map of the student model such that $\hat{y}^s_i:=M_{\theta}^s(x_i)$.
For weakly labeled data, the model mainly depends on weakly-supervised learning, and the pseudo labels $\hat{y}^{t}_i$ can be treated as a regularization term in semi-supervised training. In other words, for every batch $B$, if there are labeled data in $B$, namely $X_l \in B$, the training loss is dominated by the $L_{weak}$. Otherwise, the training loss only depends on the weighted consistency loss $L_c$. The total loss for semi-supervised training is defined as follows:



\begin{equation}
L_{semi}(\hat{y}^{s}_{i},y_{i}, y_i^f)=\left\{
\begin{array}{cl}

L_{weak}(\hat{y}^{s}_i, y_i, y_i^f) + \beta_{1} \cdot L_{c}(\hat{y}^{s}_{i}, \hat{y}^{t}_i)  \\ \label{equ6}
\beta_{2} \cdot L_{c}(\hat{y}^{s}_{i}, \hat{y}^{t}_i)  \\ 
\end{array} \right.
\end{equation}



Notably, the hyper-parameters $\beta_{1}$ and $\beta_{2}$ in equation \ref{equ6} are set to 0.1 and 0.5 respectively in this paper. For further information regarding $\beta_{1}$ and $\beta_{2}$ please refer to the supplementary material. Thus, the model is able to refine the final predicted map by considering the prior knowledge of the first rough predicted map. The overall training procedure is illustrated in Figure \ref{fig:framework_training}.

\begin{figure}[htbp]
\centering
\includegraphics[width=1\linewidth]{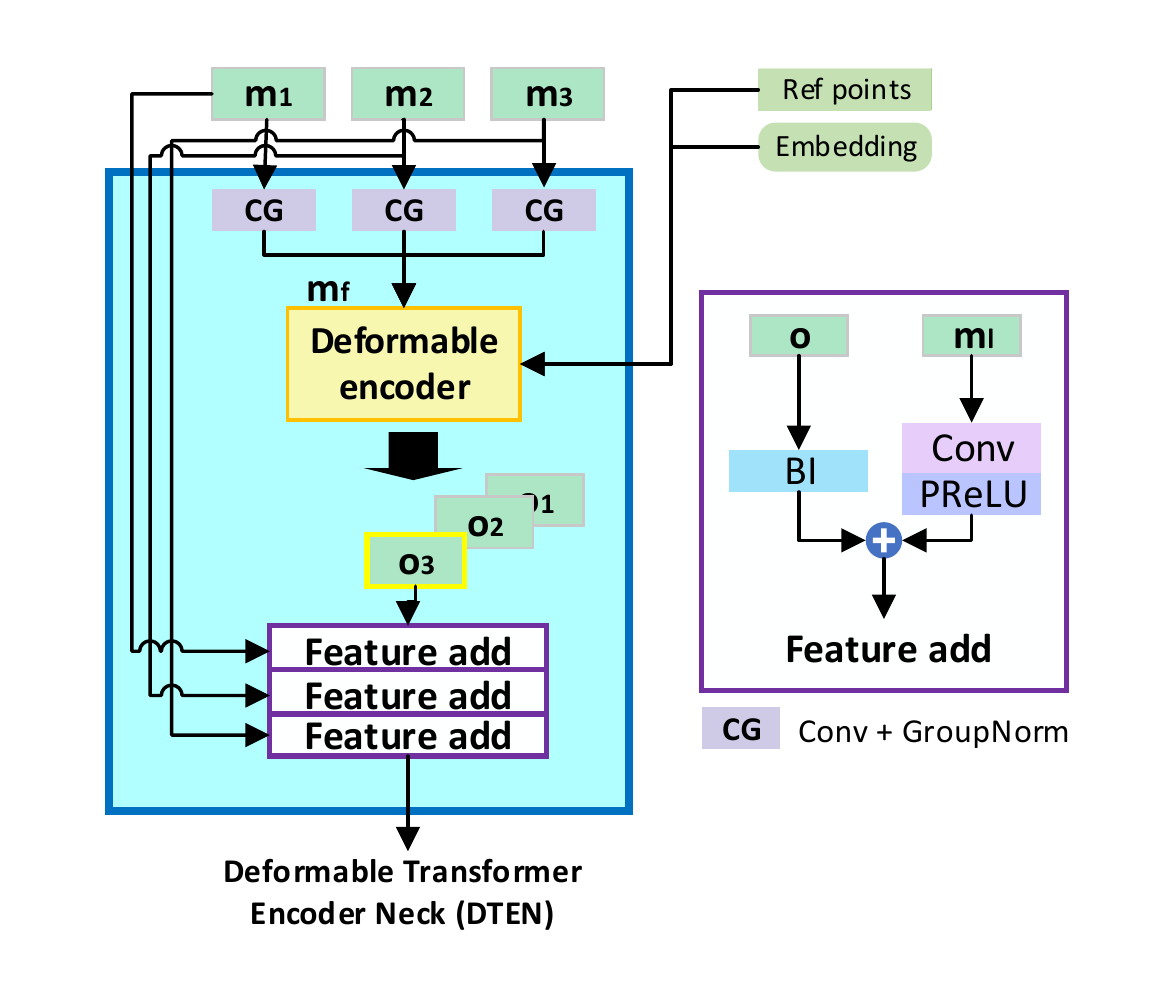}
\caption{The detailed structure of the Deformable Transformer Encoder Neck (DTEN) as described in \ref{sec:DTEN}.} 
\label{fig:deformable neck}
\end{figure}

\subsubsection{Deformable Transformer Encoder Neck}
\label{sec:DTEN}
We propose the deformable transformer encoder neck that is used after the Res2net \cite{gao2019res2net} and before the segmentation head. A detailed description of our network architecture is illustrated in Figure \ref{fig:architechture}. Its purpose is to fuse features across multiple levels and features at learned locations so that our WS-DefSegNet produces the most accurate results possible.

\paragraph{Deformable Transformer Encoder Neck (DTEN)} The structure of DTEN is illustrated in Figure \ref{fig:deformable neck}. Multi-scale feature maps $\boldsymbol{m}_l (l=1, 2, 3)$ with resolutions $H_l\times W_l$ are passed to a convolutional layer to have the same number of channels and then are normalized to have an equal contribution. Then the feature maps are flattened and concatenated to form the input feature $\boldsymbol{m_f}$. The input feature along with the pre-generated reference points and the embedding statistics \cite{DETR} are passed to the deformable encoder. The encoder outputs multi-scale enhanced feature maps $\boldsymbol{o}_l$ with resolutions the same as $\boldsymbol{m}_l$. For simplicity and sufficient details, only $\boldsymbol{o}_3$ which contains the finest features is utilized in the subsequent stacked Feature Add (FA) blocks.

\begin{table*}
\centering
  \caption{Ablation study with mDice and mIoU on five challenging datasets: ColorDB, ETIS, Kvasir, CVC-300 and ClinicDB. Upper part: the network is trained through our weak annotations. $\dagger$: denotes models trained using fully-supervised training through regular dense annotations. The best results are in \textbf{bold}.}
\label{tab:Ablation}
\resizebox{\linewidth}{!}{
\begin{tabular}{ccccccccccc}
\toprule
\multirow{2}{*}{\textbf{Method}\vspace{-6pt}
} & \multicolumn{2}{c}{\textbf{ColorDB}} & \multicolumn{2}{c}{\textbf{ETIS}} & \multicolumn{2}{c}{\textbf{Kvasir}} & \multicolumn{2}{c}{\textbf{CVC-300}} & \multicolumn{2}{c}{\textbf{ClinicDB}} \\ \cmidrule{2-11}
 & \textbf{mDice} & \textbf{mIoU} & \textbf{mDice} & \textbf{mIoU} & \textbf{mDice} & \textbf{mIoU} & \textbf{mDice} & \textbf{mIoU} & \textbf{mDice} & \textbf{mIoU}\\ \midrule

\midrule

$L_p$ & 0.327 & 0.263 & 0.218 & 0.168 & 0.555 & 0.488 & 0.240 & 0.174 & 0.479 & 0.448\\

$L_{weak}$ & 0.539 & 0.503 & 0.442 & 0.415 & 0.700 & 0.668 & 0.662 & 0.658 & 0.740 & 0.708\\

$L_{weak} + L_{c}$ & 0.604 & 0.544 & 0.501 & 0.442 & 0.730 & 0.677 & 0.729 & 0.678 & 0.771 & 0.718\\

$L_{weak}$ + DTEN & 0.609 & 0.538 & 0.541 & 0.472 & 0.728 & 0.665 & 0.754 & 0.702 & 0.772 & 0.707\\

$L_{weak}$ +DTEN+ $L_{c}$ & \textbf{0.667} & \textbf{0.588} & \textbf{0.596} & \textbf{0.517} & \textbf{0.768} & \textbf{0.709} & \textbf{0.795} & \textbf{0.728} & \textbf{0.807} & \textbf{0.746}\\

\midrule
Backbone$\dagger$ & 0.688 & 0.612 & 0.646 & 0.568 & 0.851 & 0.796 & 0.856 & 0.785 & 0.833 & 0.768\\

+DTEN$\dagger$ & \textbf{0.723} & \textbf{0.640} & \textbf{0.664} & \textbf{0.583} & \textbf{0.862} & \textbf{0.805} & \textbf{0.861} & \textbf{0.805} & \textbf{0.854} & \textbf{0.791}\\
\bottomrule
\end{tabular}
}
\end{table*}

\paragraph{Deformable Encoder} The deformable encoder \cite{DETR} enriches the input mainly by the deformable attention mechanism. It sums the selected features at learned sampling locations across multi-scales with learned attention weights. The detailed architecture of the encoder can be found in the supplementary material. The output of the encoder is then reshaped into the original resolutions, forming the enhanced multi-scale feature maps $\boldsymbol{o}_l$ as shown in Figure \ref{fig:deformable neck} for subsequent progressive feature compensation.

\paragraph{Feature Add (FA) Block}
The purpose of the FA block is to compensate the input feature map with enhanced features. The structure of a FA block is shown in Figure \ref{fig:deformable neck}.
It takes the enhanced feature map $\boldsymbol{o}$ and the original feature map $\boldsymbol{m_l}$ as inputs. The original feature map is embedded via a convolution layer and the PReLU. The enhanced map is interpolated to the same resolution as the original feature map. Three FA blocks are stacked to complement progressively the input features with the enhanced features by element-wise addition to output a more expressive feature map.

\section{Experiments}
\label{sec:experiment}

\subsection{Setup}

\paragraph{Datasets and Evaluation Metrics}
We conduct experiments on five widely used polyp datasets, namely CVC-ColonDB \cite{tajbakhsh2015colordb}, ETIS \cite{silva2014ETIS}, Kvasir \cite{jha2020kvasir}, CVC-T \cite{vazquez2017cvc-t} and CVC-ClinicDB \cite{bernal2015clinicdb}. Kvasir contains 1,000 polyp images and CVC-ClinicDB contains 612 images from 31 colonoscopy clips. The composited training images come from these two datasets and the rest of them are used for testing. The other three testing datasets are totally unseen with challenging scenarios. We follow \cite{fan2020pranet,wu2021collaborative} and employ two commonly used metrics, namely mean Dice (mDice) and mean IoU (mIoU), to evaluate the model performance for polyp segmentation.


\paragraph{Implementation Details}
Our model is implemented using Pytorch Toolbox \cite{pytorch} and trained on a GTX TITAN X GPU with a mini-batch size of 4. We adopt a 0.0005 weight decay for the Stochastic Gradient Descent (SGD) with a momentum of 0.9. For fair comparisons, both training and testing images are resized to $352\times352$, which is the same as the previous polyp segmentation methods.

\begin{table}[h]
\centering
\caption{We substitute the edge loss in \cite{zhang2020weakly} with the proposed sparse foreground loss and apply the same SOD training and testing settings as \cite{zhang2020weakly} on three SOD evaluation metrics, namely F-measure \cite{achanta2009fmeasure} (F), E-measure \cite{fan2018enhanced} (E) and mean absolute error (M).}
\label{tab:replace edge loss to foreground loss}

\begin{tabular}{cccc}
\toprule
\multirow{2}{*}{} & \textbf{Metric} & Edge & \textbf{$L_{f}$} \\
 \midrule

\multirow{3}{*}{\rotatebox{90}{ECSSD} \rotatebox{90}{}} & $F$ 
 & 0.862 & 0.854 \\
 
 & $E$ & 0.913 & 0.907 \\
 & $M$ & 0.063 & 0.063 \\ \midrule

\end{tabular}
\end{table}


\subsection{Ablation Study}

We conduct extensive experiments to analyze the merits of our proposed framework, WS-DefSegNet. 
Table \ref{tab:Ablation} ablates our framework and shows that each component, namely $L_{weak}$, $L_c$, and DTEN, boosts the segmentation performance compared to training only using $L_p$ \cite{tang2018normalized}.

\begin{table*}
\centering
  \caption{Comparisons with different semi-supervised learning methods on five challenging datasets. $L_{weak}$ refers to using the model after the weakly-supervised training without any semi-supervised training.}
\label{tab:comparisons weighted consistency loss}
\resizebox{\linewidth}{!}{
\begin{tabular}{ccccccccccc}
\toprule
\multirow{2}{*}{\textbf{Method}\vspace{-6pt}
} & \multicolumn{2}{c}{\textbf{ColorDB}} & \multicolumn{2}{c}{\textbf{ETIS}} & \multicolumn{2}{c}{\textbf{Kvasir}} & \multicolumn{2}{c}{\textbf{CVC-300}} & \multicolumn{2}{c}{\textbf{ClinicDB}} \\ \cmidrule{2-11}
 & \textbf{mDice} & \textbf{mIoU} & \textbf{mDice} & \textbf{mIoU} & \textbf{mDice} & \textbf{mIoU} & \textbf{mDice} & \textbf{mIoU} & \textbf{mDice} & \textbf{mIoU}\\ \midrule

$L_{weak}$ & 0.539 & 0.503 & 0.442 & 0.415 & 0.700 & 0.668 & 0.662 & 0.658 & 0.740 & 0.708\\
\midrule

$L_{c}$  & 0.553 & 0.508 & 0.477 & 0.431 & 0.713 & 0.665 & 0.704 & 0.678 & 0.740 & 0.693\\

$L_{weak}+L_{c}(unweighted)$  & 0.559 & 0.513 & 0.483 & 0.439 & 0.716 & 0.668 & 0.702 & 0.667 & 0.748 & 0.701\\

$L_{weak}+L_{c}(weighted)$ & 0.604 & 0.544 & 0.501 & 0.442 & 0.730 & 0.677 & 0.729 & 0.678 & 0.771 & 0.718\\

\bottomrule
\end{tabular}
}
\end{table*}

\subsubsection{Sparse Foreground Loss}
As discussed in section \ref{sec:intro}, training a model using only the partial loss $L_p$ causes a lot of false positives. Our proposed sparse foreground loss $L_f$ addresses this problem as it is shown in Figure \ref{fig:false_positives}. Compared to Figure \ref{fig:false_positives}(c), Figure \ref{fig:false_positives}(d) shows more accurate segmentation maps, which are more similar to the ground truth with fewer false positives. The benefit of our sparse foreground loss is also reflected in the overall performance as shown in Table \ref{tab:Ablation}, providing gains of up to $42.2\%$ and $48.4\%$ in terms of mDice and mIOU respectively on CVC-300.

Additionally, in contrast to the previous work \cite{zhang2020weakly}, which uses edge information and an auxiliary network to refine the segmentation maps, we do not use any extra information and networks. 
We simply use our sparse foreground loss to obtain more accurate segmentation maps and aid the model to localize objects.

In order to further demonstrate the effectiveness of our method, we conduct experiments on S-DUTS dataset \cite{zhang2020weakly} and substitute the edge loss in weakly Salient Object Detection (SOD) \cite{zhang2020weakly} with our sparse foreground loss as shown in Table \ref{tab:replace edge loss to foreground loss}. The results indicate that the proposed method can be exploited on other weakly-supervised tasks and can achieve similar performance.

\begin{figure}
\minipage{0.15\linewidth}
  \centering
  \includegraphics[width=\linewidth]{images/GT11.png}
    (a)
\endminipage\hfill
\minipage{0.15\linewidth}
  \centering
  \includegraphics[width=\linewidth]{images/baseline11.png}
    (b) 
\endminipage\hfill
\minipage{0.15\linewidth}
  \centering
  \includegraphics[width=\linewidth]{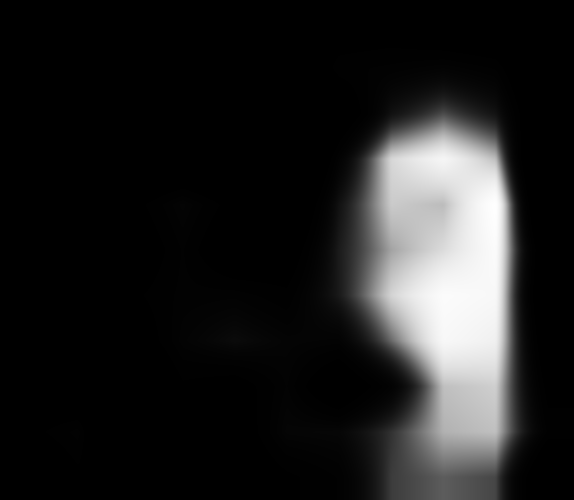}
    (c)
\endminipage\hfill
\minipage{0.15\linewidth}
  \centering
  \includegraphics[width=\linewidth]{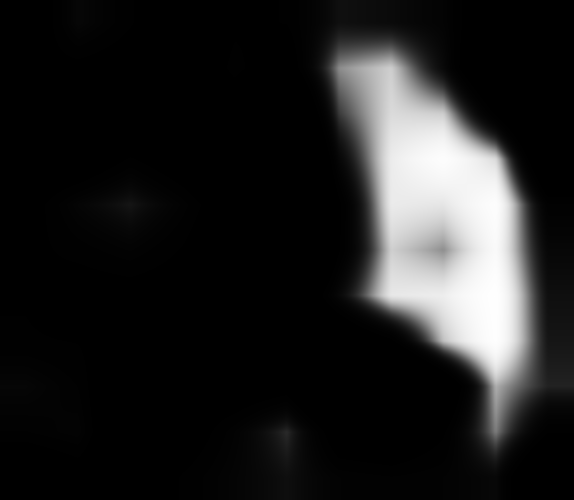}
    (d)
\endminipage\hfill
\minipage{0.15\linewidth}
  \centering
  \includegraphics[width=\linewidth]{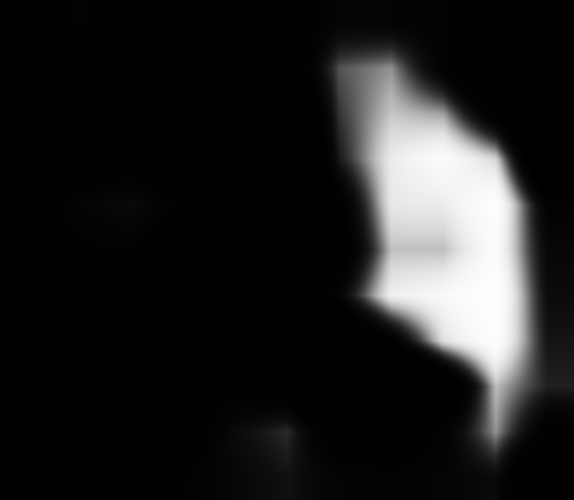}
    (e)
\endminipage\hfill
\minipage{0.15\linewidth}
  \centering
  \includegraphics[width=\linewidth]{images/+semi11.png}
    (f)
\endminipage\hfill

\caption{Difference between predictions of the two identical backbones with the same training settings and different semi-supervised methods. (a) Ground truth. (b) Predicted segmentation map of the first model. (c) Predicted segmentation map of the second model. (d) Only pseudo labels for semi-training. (e) $L_{semi}$ without $\beta$ for semi-training. (f) Ours.}
\label{fig: weak issues}
\end{figure}

\subsubsection{Batch-wise Weighted Consistency Loss}
We add the proposed batch-wise weighted consistency loss $L_{c}$ to the baseline $L_{weak}$ in Table \ref{tab:Ablation} for the semi-supervised training. The experimental results show that this method can increase the segmentation accuracy on both mDice and mIoU across all testing datasets. It can also be observed in Figure \ref{fig:false_positives}(e) that $L_c$ eliminates the false positive pixels next to the polyp, and also improves the predicted segmentation maps compared with Figure \ref{fig:false_positives}(d).

We show the superiority of the proposed weighted consistency loss in Table \ref{tab:comparisons weighted consistency loss}. Apparently, without the aid of weights $\beta_1$ and $\beta_2$, $L_{weak}$ contributes minor improvement compared to only training on pseudo labels during semi-supervised training. Our weighted consistency loss addresses the inconsistent issue caused by weak supervision (Figure \ref{fig: weak issues}(b) and (c)) by taking full advantage of the two predicted maps for semi-supervised training. When compared to using only pseudo-labels or using $L_{semi}$ without weights $\beta_1$ and $\beta_2$, it can be seen from Figure \ref{fig: weak issues}(d), (e) and (f) our proposed solution provides a more accurate segmentation map with refined boundaries. Both Table \ref{tab:comparisons weighted consistency loss} and Figure \ref{fig: weak issues} demonstrate the effectiveness of our proposed method.

\subsubsection{DTEN}
To investigate whether the proposed DTEN benefits polyp segmentation, we compare the results with and without DTEN under different training regimes as shown in Table \ref{tab:Ablation}. Regarding the weakly- and semi-supervised training part, DTEN provides significant performance increase under all metrics and on all datasets when compared to using only our proposed loss functions. 
In the fully supervised training section, applying DTEN on top of the Res2net also enhances the performance. 
These results indicate the effectiveness of the proposed structure and demonstrate the importance of enhancing features for accurate segmentation. 

\begin{table*}
\centering
  \caption{Evaluation results of different methods on five datasets.*uses semi-supevised training. Ours: denotes our method that is trained using weakly- and semi-supervised training.}
\label{tab:comparisons with sotas}
\resizebox{\linewidth}{!}{
\begin{tabular}{ccccccccccccc}
\toprule
\multirow{2}{*}{\textbf{Method}\vspace{-6pt}
}  & \multirow{2}{*}{\textbf{Average}
}& \multicolumn{2}{c}{\textbf{ColorDB}} & \multicolumn{2}{c}{\textbf{ETIS}} & \multicolumn{2}{c}{\textbf{Kvasir}} & \multicolumn{2}{c}{\textbf{CVC-300}} & \multicolumn{2}{c}{\textbf{ClinicDB}} \\ \cmidrule{3-12}
 & \textbf{Labeled Pixels}& \textbf{mDice} & \textbf{mIoU} & \textbf{mDice} & \textbf{mIoU} & \textbf{mDice} & \textbf{mIoU} & \textbf{mDice} & \textbf{mIoU} & \textbf{mDice} & \textbf{mIoU}\\ \midrule

\midrule

U-Net(MICCAI’15)\cite{ronneberger2015unet}  & $13.4\%$ & 0.512 & 0.444 & 0.398 & 0.335  & 0.818 & 0.746 & 0.710 & 0.627 & 0.823 & 0.755\\

U-Net++(TMI’19)\cite{zhou2018unet++}  & $13.4\%$ & 0.483 & 0.410 & 0.401 & 0.344 & 0.821 & 0.743 & 0.707 & 0.624 & 0.794 & 0.729\\

ResUNet++(ISM'19)\cite{jha2019resunet++}  & $13.4\%$ & - & - & - & - & 0.813 & 0.793 & - & - & 0.796 & 0.796\\

SFA(MICCAI’19)\cite{fang2019sfa}  & $13.4\%$ & 0.469 & 0.347 & 0.297 & 0.217 & 0.723 & 0.611 & 0.467 & 0.329 & 0.700 & 0.607\\

PraNet(MICCAI'20)\cite{fan2020pranet}  & $13.4\%$ & 0.709 & 0.640 & 0.628 & 0.567 & 0.898 & 0.840 & 0.871 & 0.797 & 0.899 & 0.849\\

CAL(ICCV'21)*\cite{wu2021collaborative}   & $4.0\%$ & - & - & - & - & 0.810 & 0.716 & - & - & 0.893 & 0.826\\

Ours  & $1.9\%$ & 0.667 & 0.588 & 0.596 & 0.517 & 0.768 & 0.709 & 0.795 & 0.728 & 0.807 & 0.746\\

\bottomrule
\end{tabular}
}
\end{table*}

\begin{table*}
\centering
  \caption{Fine-tuning results with mDice and mIoU on five challenging datasets for different state-of-the-art approaches. $\dagger$: denotes models trained using fully supervised training through regular dense annotations. The best results are in \textbf{bold}.}
\label{tab:ddifferent_networks}
\resizebox{\linewidth}{!}{
\begin{tabular}{ccccccccccc}
\toprule
\multirow{2}{*}{\textbf{Method}\vspace{-6pt}
} & \multicolumn{2}{c}{\textbf{ColorDB}} & \multicolumn{2}{c}{\textbf{ETIS}} & \multicolumn{2}{c}{\textbf{Kvasir}} & \multicolumn{2}{c}{\textbf{CVC-300}} & \multicolumn{2}{c}{\textbf{ClinicDB}} \\ \cmidrule{2-11}
 & \textbf{mDice} & \textbf{mIoU} & \textbf{mDice} & \textbf{mIoU} & \textbf{mDice} & \textbf{mIoU} & \textbf{mDice} & \textbf{mIoU} & \textbf{mDice} & \textbf{mIoU}\\ \midrule

\midrule
Poolnet(pretrained)\cite{poolnet} & 0.159 & 0.103 & 0.086 & 0.057 & 0.455 & 0.361 & 0.135 & 0.084 & 0.240 & 0.171\\

Poolnet$\dagger$  & 0.439 & 0.403 & 0.330 & 0.327 & 0.774 & \textbf{0.743} & 0.543 & 0.528 & 0.629 & 0.622\\

Ours($L_{weak}$) & 0.576 & 0.508 & 0.426 & 0.383 & 0.743 & 0.682 & 0.722 & 0.649 & 0.763 & 0.701\\

Ours($L_{weak} + L_{c}$) & \textbf{0.583} & \textbf{0.508} & \textbf{0.459} & \textbf{0.415} & \textbf{0.776} & 0.708 & \textbf{0.755} & \textbf{0.676} & \textbf{0.782} & \textbf{0.721}\\


\midrule

A2dele(pretrained)\cite{a2dele} & 0.219 & 0.153 & 0.225 & 0.161 & 0.470 & 0.352 & 0.359 & 0.271 & 0.287 & 0.195\\

A2dele$\dagger$  & 0.450 & 0.461 & 0.378 & 0.406 & \textbf{0.706} & \textbf{0.713} & 0.666 & 0.718 & 0.588 & 0.633\\

Ours($L_{weak}$) & 0.487 & 0.500 & 0.413 & 0.440 & 0.610 & 0.605 & 0.660 & 0.702 & 0.579 & 0.601\\

Ours($L_{weak} + L_{c}$)  & \textbf{0.509} & \textbf{0.511} & \textbf{0.449} & \textbf{0.457} & 0.662 & 0.645 & \textbf{0.695} & \textbf{0.728} & \textbf{0.623} & \textbf{0.636}\\





\bottomrule
\end{tabular}
}

\end{table*}

\begin{table}
\centering
  \caption{Quantitative results with mDice and mIoU on DiNO.}
\label{tab:dino}
\resizebox{\linewidth}{!}{
\begin{tabular}{ccccc}
\toprule
\multirow{2}{*}{\textbf{Method}\vspace{-6pt}
} & \multicolumn{2}{c}{\textbf{ColorDB}} & \multicolumn{2}{c}{\textbf{ClinicDB}} \\ \cmidrule{2-5}
 & \textbf{mDice} & \textbf{mIoU} & \textbf{mDice} & \textbf{mIoU} \\ \midrule

\midrule

DiNO+$L_{weak}$ & 0.577 & 0.489 & 0.756 & 0.670\\

DiNO+$L_{weak} + L_{c}$  & 0.623 & 0.527  & 0.821 & 0.747\\

\bottomrule
\end{tabular}
}

\end{table}

\subsection{Comparison with the state-of-the-arts}

To further validate our proposed framework, we compare it with other state-of-the-art methods, namely, U-Net \cite{ronneberger2015unet}, U-Net++ \cite{zhou2018unet++}, ResUNet++ \cite{jha2019resunet++}, SFA \cite{fang2019sfa}, PraNet \cite{fan2020pranet} and CAL \cite{wu2021collaborative} on five challenging polyp testing datasets. We directly report the results provided by each work. It should be noted that we are the only ones using weakly annotated images. Our results show that we can compete and even surpass methods that were trained in a fully supervised way as seen in Table \ref{tab:comparisons with sotas}. Also, we obtain competitive results compared to \cite{wu2021collaborative} which is the only other method that uses semi-supervised training. However, in contrast to our framework, \cite{wu2021collaborative} uses pixel-wise annotated images while we only use weakly-annotated images. Also, our method uses less than half of the averaged labeled pixels that \cite{wu2021collaborative} uses.

It is also worth noting that other state-of-the-art methods \cite{ronneberger2015unet, zhou2018unet++, jha2019resunet++, fang2019sfa,wu2021collaborative} may suffer from overfitting issues because they only obtain high performance on Kvasir and ClinicDB. Compared to them, ours achieves satisfactory performance on all five testing datasets. The results in Table \ref{tab:comparisons with sotas} demonstrate the superior generalization ability of our framework.

\subsection{Transfer Learning on Other Networks}
In order to investigate the transferability of our method, we leverage our framework to adapt other networks that were trained on different tasks. First of all, we use two pre-trained SOD detectors, the RGB-trained Poolnet \cite{poolnet} and the RGB-D trained A2dele \cite{a2dele}, and show that we can fine-tune them successfully using our novel loss functions $L_{weak}$ and $L_c$ as shown in Table \ref{tab:ddifferent_networks}. The baseline results show how each method performs without any adaptation. Impressively, simply fine-tuning both A2dele and Poolnet using our proposed loss functions outperforms fine-tuning in a fully supervised way. These results highlight the transfer learning ability of our framework and its potential to be used for different networks. Similarly to Table \ref{tab:Ablation}, it can be seen that each of our proposed loss functions provides a significant performance increase.

Furthermore, the generality of our method can be seen (Table \ref{tab:dino}) beyond convolutional-based backbones. Using our framework we fine-tune a transformer-based backbone, DiNO \cite{dino}, and a convolutional-based segmentation head surpassing the performance of other fully supervised polyp segmentation methods.

\section{Conclusion}
\label{sec:conclusion}

In this paper, we propose a novel framework WS-DefSegNet for weakly- and semi-supervised polyp segmentation. We create a weakly annotated polyp dataset (W-Polyp) by simply drawing sketches. This annotating method provides an efficient way for physicians to avoid manual labour. 

We propose a sparse foreground loss that suppresses false positives. Furthermore, we propose a batch-wise weighted consistency loss to exploit two inconsistent segmentation maps caused by weak supervision during semi-supervised training. Also, we design a deformable transformer encoder neck (DTEN) as a way to enhance features before the segmentation head further improving performance.

Extensive experiments are conducted on five challenging datasets to demonstrate that each proposed component improves the segmentation accuracy and that our framework can even surpass the performance of some state-of-the-art methods trained in a fully supervised way.


\newpage
{\small
\bibliographystyle{ieee_fullname}
\bibliography{egbib}
}

\newpage

\appendix
\begin{center}
\textbf{\Large Supplementary material}
\end{center}


\section{Deformable transformer encoder}
\begin{figure}[h]
\centering
\includegraphics[width=1\linewidth]{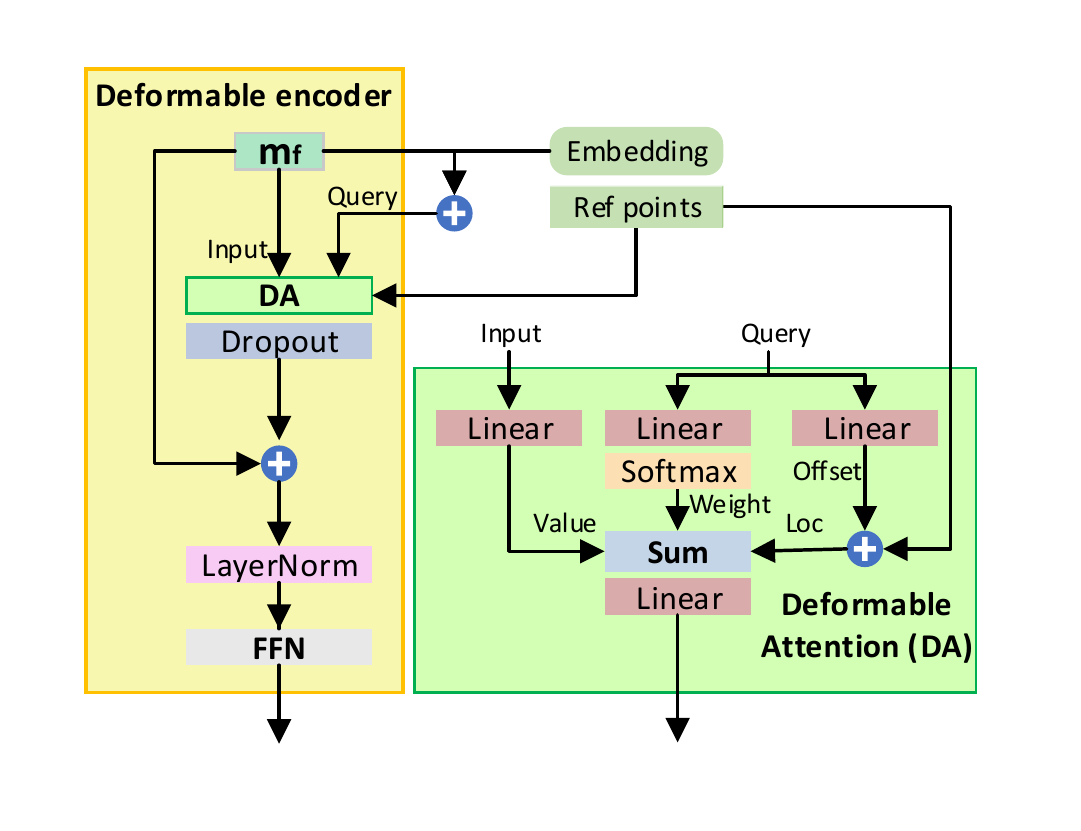}

\caption{The detailed structure of a single encoder in deformable vision transformers. The encoder enables the aggregation of useful features at learned locations with learnt significance across levels.}
\label{fig:encoder}
\end{figure}

The deformable encoder \cite{DETR} enriches the input mainly by the Deformable Attention (DA) module and the Feed-Forward Network (FFN). The detailed architecture can be seen in Figure \ref{fig:encoder}. The DA module sums the selected features at deformable sampling locations across multi-scales with learned attention weights. The ourput of this module is passed through the FFN. 

Suppose the encoder takes as inputs the flattened feature map $\boldsymbol{m_f}\in\mathbb{R}^{C\times N_{in}} (C$: number of channels, $N_{in}=\sum_l H_l W_l)$, positional and level embedding information $\boldsymbol{E}\in\mathbb{R}^{C\times N_{in}}$ and reference points $\boldsymbol{P}\in\mathbb{R}^{N_{in}\times 2}$. The output of the deformable attention layer is formulated as:
\begin{equation}
\centering
    \boldsymbol{O_{DA}}=f(\boldsymbol{O_s})
\end{equation}
where $f$ is the linear layer and $\boldsymbol{O_s}$ is the weighted summation:
\begin{equation}
    \centering
    \boldsymbol{O}_s=\sum_{l,p} \boldsymbol{W}_{nhlp}\boldsymbol{V}_{nh}(\boldsymbol{P}_n+\Delta \boldsymbol{P}_{nhlp})\xrightarrow[]{} [C,N_{in}]
\end{equation}
where $n$, $h$, $l$ and $p$ index the pixel in the flattened feature map, attention head, feature level and sampling point respectively. The rightarrow $\xrightarrow[]{}$ represents reshaping to the dimensions in the brackets.
The value feature $\boldsymbol{V}$, the predicted sampling offsets $\Delta \boldsymbol{P}$ and attention weights $\boldsymbol{W}$ are defined as: 
\begin{equation}
\centering
\boldsymbol{V}=f(\boldsymbol{m}\xrightarrow[]{}[N_{in},N_h,C/N_h])
\end{equation}
\begin{equation}
\centering
\Delta \boldsymbol{P}=f(\boldsymbol{Q}) \xrightarrow[]{}[N_{in},N_h,N_l,N_p,2]
\end{equation}
\begin{equation}
\begin{split}
\boldsymbol{W}=Softmax(f(\boldsymbol{Q}) \xrightarrow[]{}[N_{in},N_h,N_l N_p])\\ \xrightarrow[]{}[N_{in},N_h,N_l,N_p]
\end{split}
\end{equation}
where the query feature $\boldsymbol{Q}$ is the element-wise addition:
\begin{equation}
\centering
    \boldsymbol{Q}=\boldsymbol{m}+\boldsymbol{E}
\end{equation}
It should be noted that $\boldsymbol{W}$ is normalized in the last dimension to provide weights that sum up to 1.
The encoder finally outputs $\boldsymbol{O}\in \mathbb{R}^{C\times N_{in}}$ as:
\begin{equation}
    \boldsymbol{O}=FFN(LN(Dropout(\boldsymbol{O}_{DA})+\boldsymbol{m})) 
\end{equation}
where $FFN$ and $LN$ are short for Feed-Forward Network and Layer Normalization layer respectively.

\section{Dataset explanations}
To acquire the weak annotations of polyps, we manually annotate the simple sketches with the help of the PaintTool SAI, which is a painting tool for drawing. Annotators are asked to relabel the dataset according to their first impressions without a fixed drawing style. These simple sketches only cost 2 seconds to label an image.

\begin{figure*}
\minipage{\textwidth}
  \centering
  \includegraphics[width=0.9\linewidth]{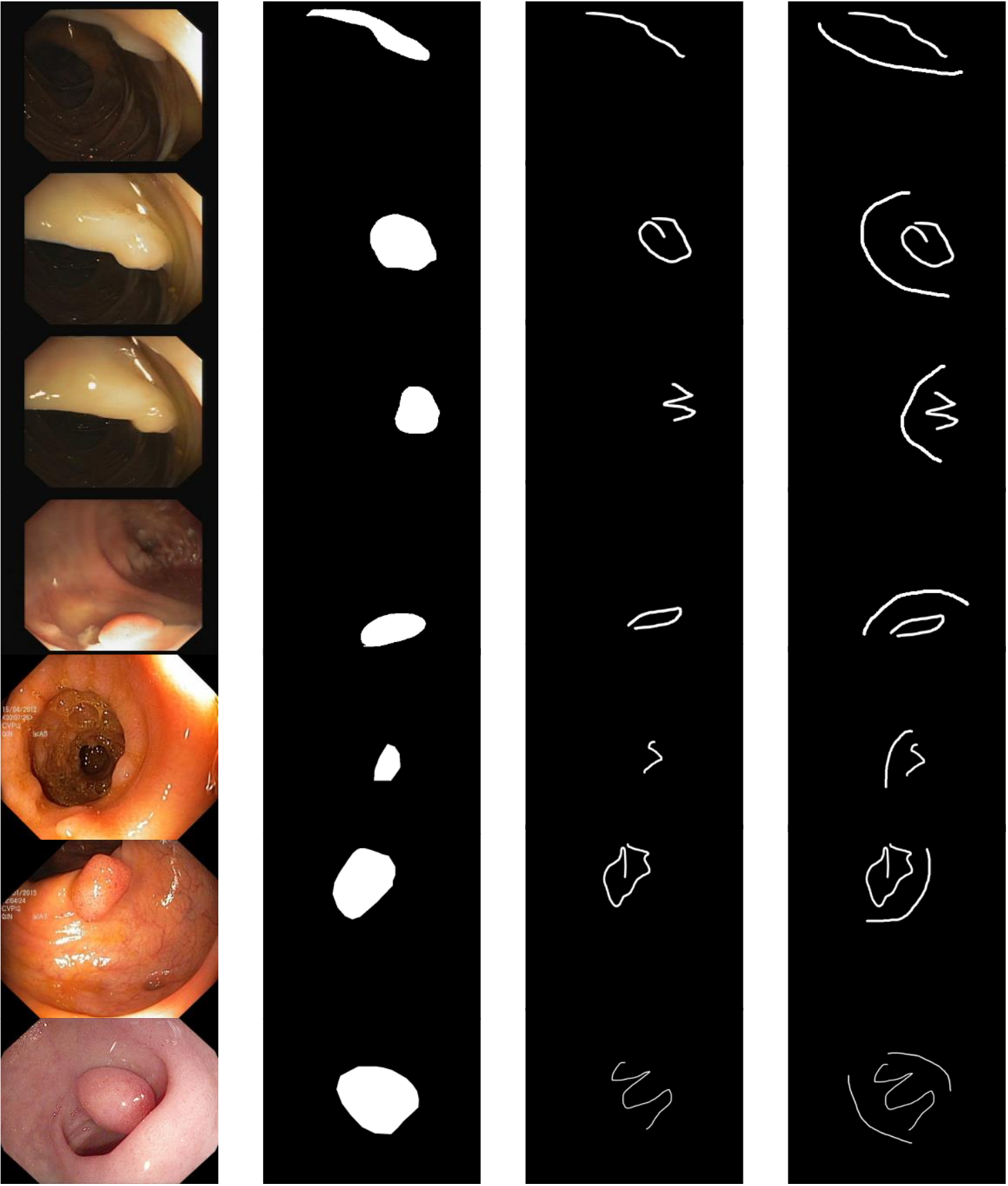}
\endminipage\hfill
\caption{Training samples.}
\label{fig:weak annotations in the training dataset}
\end{figure*}

More visualizations of our annotated dataset can be seen in Figure \ref{fig:weak annotations in the training dataset}. Column 1 shows the original image, column 2 shows the original ground truth segmentation map, column 3 shows only the foreground annotation, and column 4 shows both the background and foreground annotations. Only the annotations shown in the last 2 columns were used in training our model.

\begin{table*}[ht]
\centering
  \caption{Comparisons with different $\alpha$ in weakly-supervised training.}
\label{tab:comparisons alpha}
\resizebox{\linewidth}{!}{
\begin{tabular}{ccccccccccc}
\toprule
\multirow{2}{*}{\textbf{$\alpha$}\vspace{-6pt}
} & \multicolumn{2}{c}{\textbf{ColorDB}} & \multicolumn{2}{c}{\textbf{ETIS}} & \multicolumn{2}{c}{\textbf{Kvasir}} & \multicolumn{2}{c}{\textbf{CVC-300}} & \multicolumn{2}{c}{\textbf{ClinicDB}} \\ \cmidrule{2-11}
 & \textbf{mDice} & \textbf{mIoU} & \textbf{mDice} & \textbf{mIoU} & \textbf{mDice} & \textbf{mIoU} & \textbf{mDice} & \textbf{mIoU} & \textbf{mDice} & \textbf{mIoU}\\ \midrule

\midrule
0 & 0.327 & 0.263 & 0.218 & 0.168 & 0.555 & 0.488 & 0.240 & 0.174 & 0.479 & 0.448\\

0.5 & 0.539 & 0.503 & 0.442 & 0.415 & 0.700 & 0.668 & 0.662 & 0.658 & 0.740 & 0.708\\

1 & 0.124 & 0.089 & 0.064 & 0.026 & 0.209 & 0.133 & 0.060 & 0.029 & 0.126 & 0.082\\

\bottomrule
\end{tabular}
}

\end{table*}

\begin{table*}
\centering
  \caption{Comparisons with different combinations of $\beta_1$ and $\beta_2$ in semi-supervised training. 'Baseline' represents the performance of the model trained only with $L_{weak}$ using equation \ref{eq:l_weak}.}
\label{tab:comparisons beta}
\resizebox{\linewidth}{!}{
\begin{tabular}{ccccccccccc}
\toprule
\textbf{($\beta_1$, $\beta_2$)} & \multicolumn{2}{c}{\textbf{ColorDB}} & \multicolumn{2}{c}{\textbf{ETIS}} & \multicolumn{2}{c}{\textbf{Kvasir}} & \multicolumn{2}{c}{\textbf{CVC-300}} & \multicolumn{2}{c}{\textbf{ClinicDB}} \\ \cmidrule{2-11}
\textbf{} & \textbf{mDice} & \textbf{mIoU} & \textbf{mDice} & \textbf{mIoU} & \textbf{mDice} & \textbf{mIoU} & \textbf{mDice} & \textbf{mIoU} & \textbf{mDice} & \textbf{mIoU}\\ \midrule

\midrule
Baseline & 0.539 & 0.503 & 0.442 & 0.415 & 0.700 & 0.668 & 0.662 & 0.658 & 0.740 & 0.708\\

$(0.5,0.5)$  & 0.559 & 0.513 & 0.483 & 0.439 & 0.716 & 0.668 & 0.702 & 0.667 & 0.748 & 0.701\\

$(0.3,0.5)$  & 0.579 & 0.527 & 0.497 & 0.444 & 0.718 & 0.668 & 0.722 & 0.692 & 0.760 & 0.716\\

$(0.1,0.5)$  & 0.604 & 0.544 & 0.501 & 0.442 & 0.730 & 0.677 & 0.729 & 0.678 & 0.771 & 0.718\\

$(0.0,0.5)$  & 0.582 & 0.511 & 0.424 & 0.359 & 0.759 & 0.690 & 0.648 & 0.585 & 0.756 & 0.690\\

\bottomrule
\end{tabular}
}

\end{table*}

\section{Visualizations}

\begin{figure*}

\minipage{\textwidth}
  \centering
  \includegraphics[width=\linewidth]{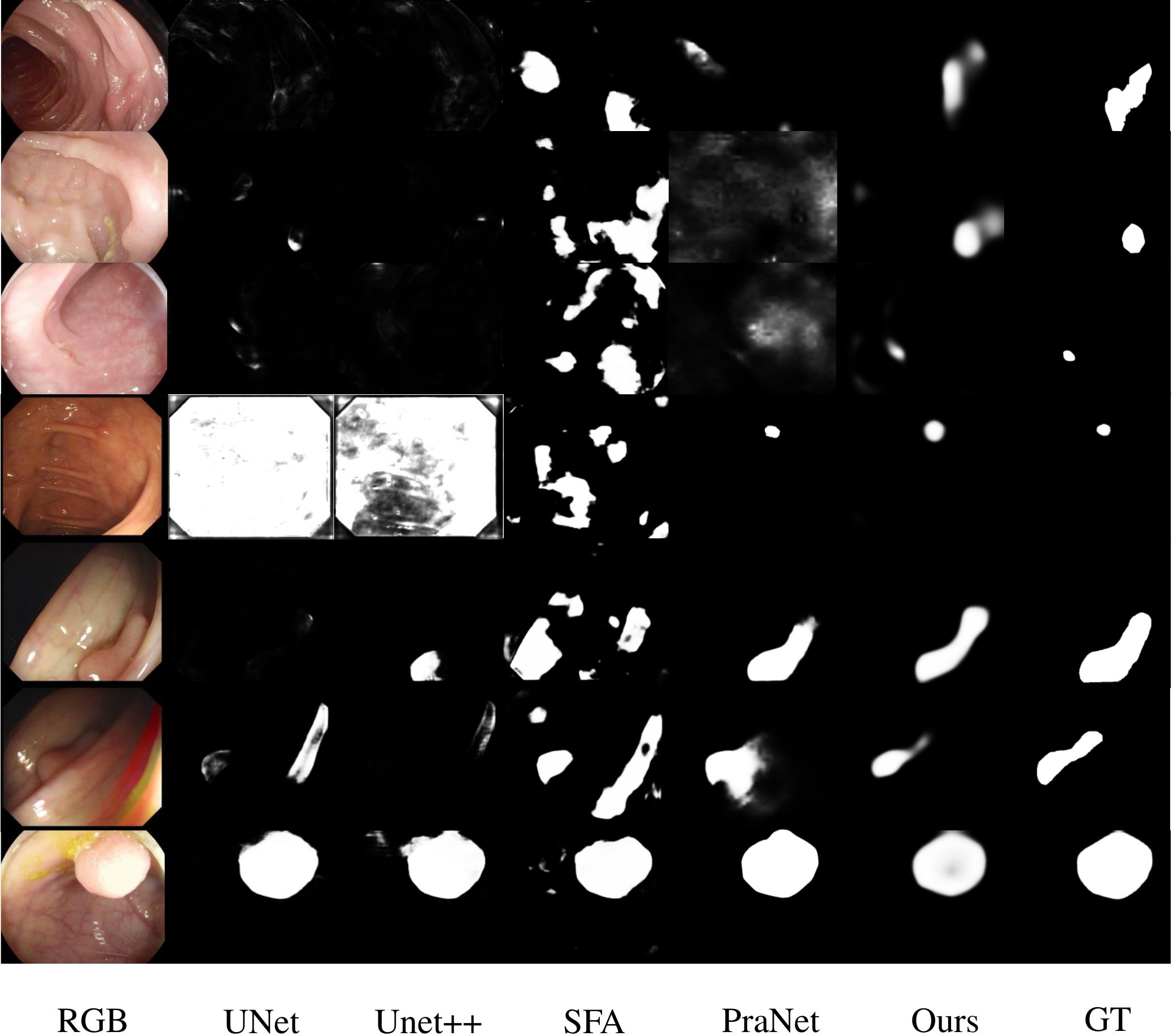}
\endminipage\hfill
\caption{Comparisons with other state of the art methods.}
\label{fig:sota_comparisons_appendix}
\end{figure*}

Figure \ref{fig:sota_comparisons_appendix} shows qualitative results between our method and other state-of-the-art methods. It should be noted that all the other methods shown were trained in a fully supervised way. Impressively, in some cases such as in rows 1, 2 and 3, the fully supervised methods completely fail while our method manages to recover the main polyp part. In order to provide fair visualizations and avoid cherry-picking we also provided more cases where other methods such as PraNet \cite{fan2020pranet} perform better such as rows 4, 5 and 7. However, our method still outperforms all other fully supervised methods beyond Pranet in these qualitative visualizations. In other words, visual maps in Figure \ref{fig:sota_comparisons_appendix} demonstrate that the proposed method has a better generalization ability that can achieve satisfactory detection results in different scenarios.

\begin{figure*}

\minipage{\textwidth}
  \centering
  \includegraphics[width=0.8\linewidth]{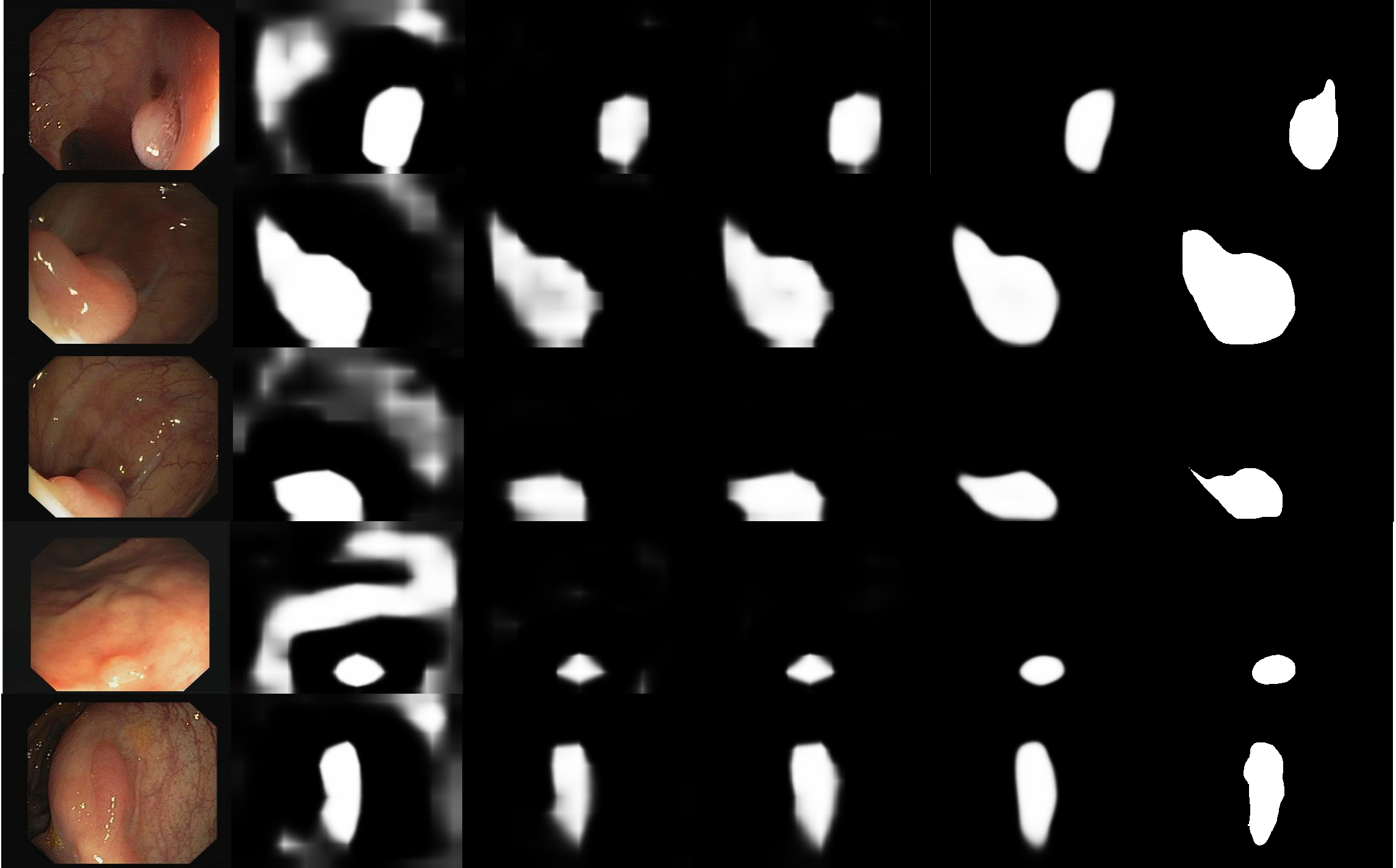}
\endminipage\hfill
\caption{ClinicDB}
\label{fig:loss_comparison_clinic_ClinicDB}
\end{figure*}

\begin{figure*}

\minipage{\textwidth}
  \centering
  \includegraphics[width=0.8\linewidth]{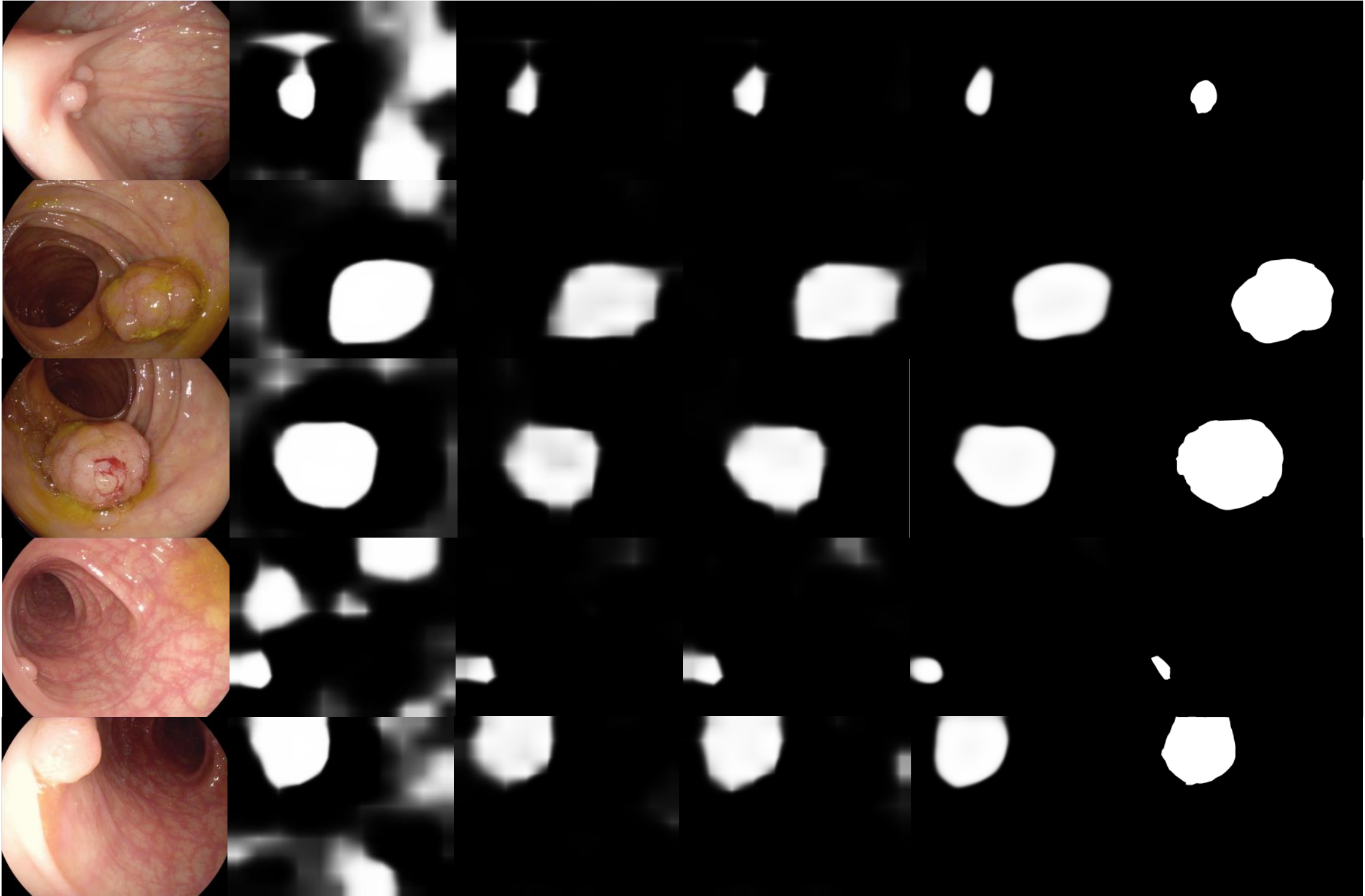}
\endminipage\hfill
\caption{ETIS}
\label{fig:loss_comparison_clinic_ET}
\end{figure*}

\begin{figure*}

\minipage{\textwidth}
  \centering
  \includegraphics[width=0.8\linewidth]{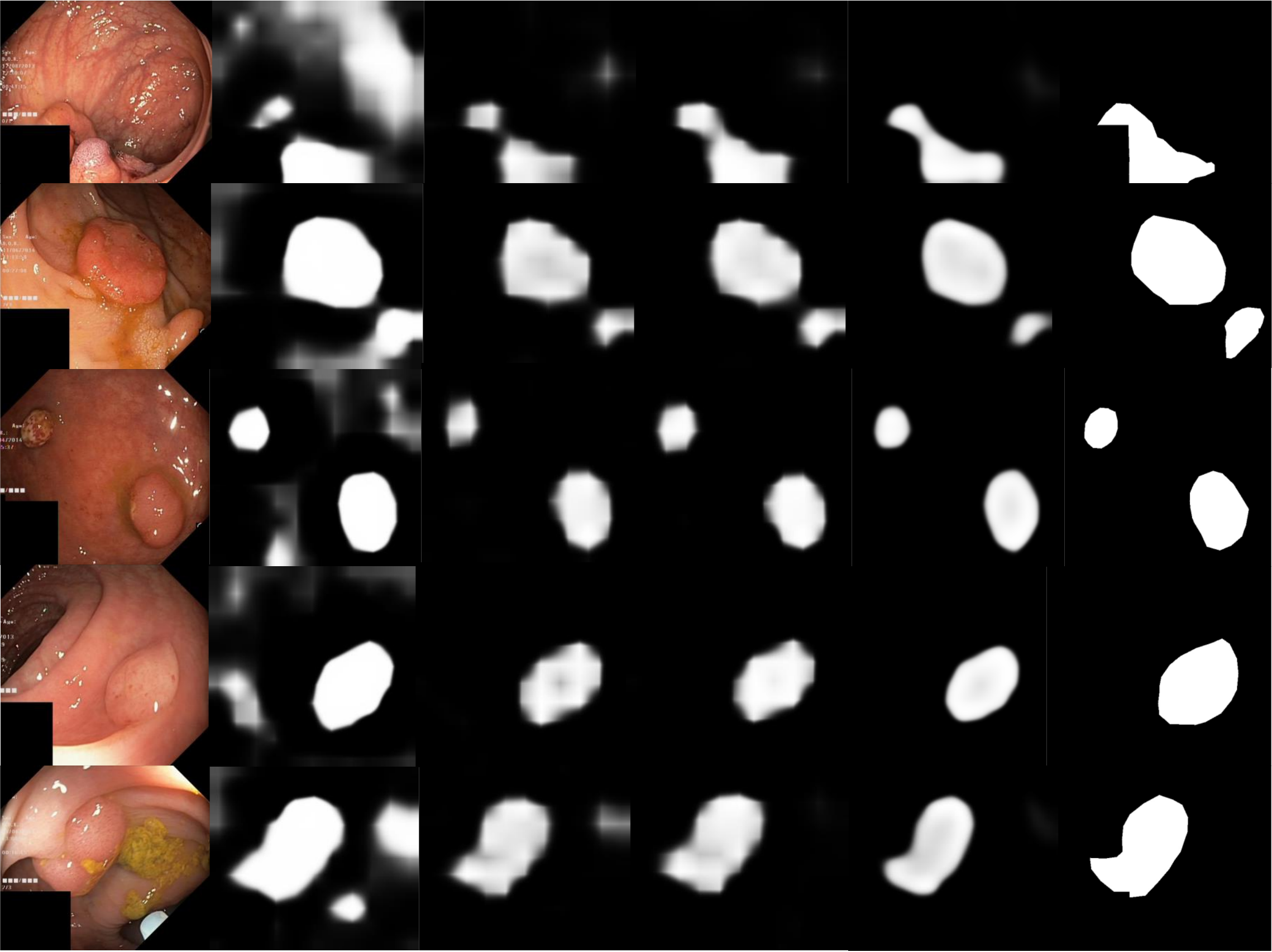}
\endminipage\hfill
\caption{Kavsir}
\label{fig:loss_comparison_clinic_Kavsir}
\end{figure*}

\begin{figure*}

\minipage{\textwidth}
  \centering
  \includegraphics[width=0.8\linewidth]{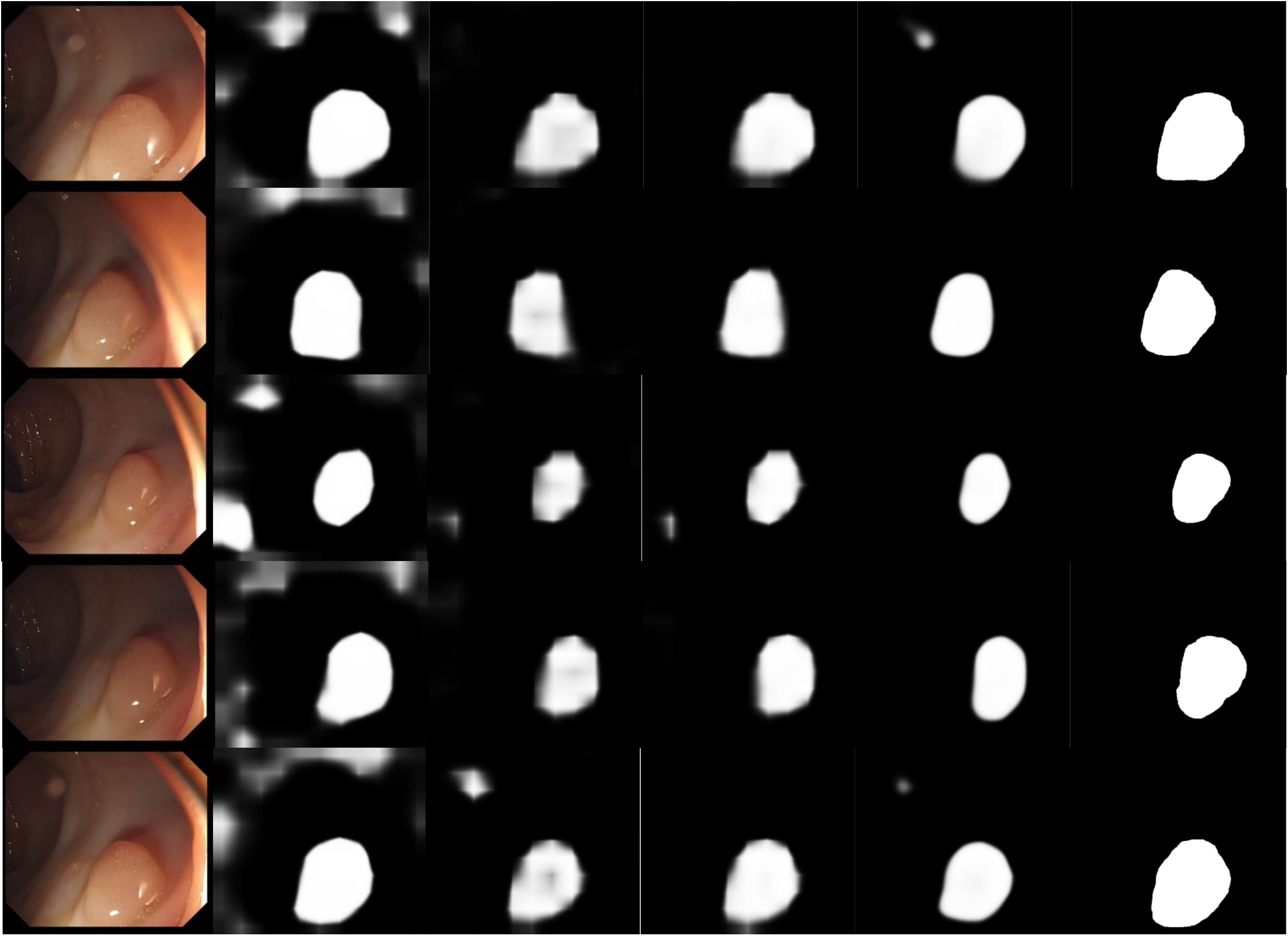}
\endminipage\hfill
\caption{ColorDB}
\label{fig:loss_comparison_clinic_ColorDB}
\end{figure*}

Lastly Figures \ref{fig:loss_comparison_clinic_ClinicDB}, \ref{fig:loss_comparison_clinic_ET}, \ref{fig:loss_comparison_clinic_Kavsir}, and \ref{fig:loss_comparison_clinic_ColorDB} show more qualitative examples ablating our method in a similar way to Figure \ref{fig:false_positives}. Column 1 shows the original RGB image, column 2 shows the prediction when trained only with $L_p$, column 3 shows the predictions when trained with sparse foreground loss, column 4 shows the predictions when trained with $L_{semi}$, column 5 shows the predictions when trained using DTEN and column 6 shows the original ground truth segmentation maps. It is evident that each proposed idea of our method provides performance improvement evident from these visualizations.

\section{Hyperparameter optimization}

In order to find proper $\alpha$ (equation \ref{eq:l_weak}), $\beta_1$ and $\beta_2$ (equation \ref{equ6}) for our training regime, we carried out hyperparameter optimization. We investigated the performance of our regime on five polyp datasets with different hyperparameter settings as presented in Tables \ref{tab:comparisons alpha} and \ref{tab:comparisons beta}. Results show that the most accurate segmentation is achieved on all datasets with $\alpha=0.5$. For $(\beta_1$, $\beta_2)$, the combination of $(0.1, 0.5)$ produces the best results on three out of the five datasets. To generalize the regime, we use the aforementioned settings for most robust and accurate segmentation.



		 
		
		
		
 
 

\clearpage
\begin{table*}[h]
\centering
  \caption{Notations lookup table.}
\label{tab:notations}
\begin{tabular}{ll} 
\toprule
Notation                                   & Description                                             \\ 
\midrule
$X$                                        & Whole dataset                                           \\
$X_l$                                      & Weakly-annotated
  subset                               \\
$Y_l$                                      & Weakly-annotated
  subset’s ground truth                \\
$X_u$                                      & Unlabeled
  subset                                      \\
$\theta$                                   & Learnable
  parameters                                  \\
$M_\theta$                                 & Model                                                   \\
$M_\theta^t$                               & Teacher model                                           \\
$M_\theta^s$                               & Student
  model                                         \\
$x_i$                                      & The $i$th image from $X$                                                      \\
${\hat{y}}_i$                              & Predicted segmentation map of~$x_i$                     \\
$y_i$                                      & Ground truth map of~$x_i$                               \\
$y_i^f$                                    & The ground truth map of~$x_i$ with only the foreground annotation                                                   \\
$L_p$                                      & Partial cross-entropy loss                            \\
$L_f$                                      & Sparse foreground loss                                       \\
$L_{weak}$                                 & Total loss for weakly-supervised learning             \\
$\alpha$                                   & Weight
  of the foreground loss                         \\
${\hat{y}}_i^t$, ${\hat{y}}_i^s$           &  The prediction of the teacher and student models with input~$x_i$               \\
$B$                                        & Batch                                                   \\
$B_l$                                      & The batch of labeled samples                                             \\
$B_l^f$                                    & The batch of foreground annotations                           \\
$L_{semi}$                                 & Total loss for semi-supervised learning in each~$B$     \\
$\beta_1$, $\beta_2$                       & Weights of $L_c$ in $L_{semi}$                                         \\
$\boldsymbol{m}$\textbf{}                  & Feature
  map output by the last stage of the backbone  \\
$l$                                        & Index of
  the feature level                            \\
$n$                                        & Index of the pixel in~$\boldsymbol{m_f}$                \\
$h$                                        & Index of
  the attention head                           \\
$p$                                        & Index of
  the sampling point                           \\
$\boldsymbol{m}_l$                         & Feature map at $l$-th level                             \\
$H_l$, $W_l$                               & Height and width of~$\boldsymbol{m}_l$                            \\
$W_l$                                      & Width of~$\boldsymbol{m}_l$                             \\
$\boldsymbol{m_f}$                         & Feature
  map after concatenation and flatten           \\
$\boldsymbol{o}_l$                         & Output feature map by the encoder at $l$-th level       \\
$C$                                        & Number of
  channels                                    \\
$N_{in}$                                   & Number of pixels in~$\boldsymbol{m_f}$                  \\
$N_h$                                      & Number of
  attention heads                             \\
$N_l$                                      & Number of
  levels                                      \\
$N_p$                                      & Number of
  sampling points                             \\
$\mathbb{R}$                               & Real
  number                                           \\
$\boldsymbol{P}$                           & Reference
  points                                      \\
$\boldsymbol{E}$ \textbf{}                 & Position
  and level embedding information              \\
$\boldsymbol{O}$ \textbf{}                 & Output
  of the encoder                                 \\
$f$                                        & Linear
  layer                                          \\
$\boldsymbol{W}$                           & Attention
  weights                                     \\
$\boldsymbol{V}$ \textbf{}                 & Value
  tensor                                          \\
$\boldsymbol{\Delta P}$ \textbf{\textit{}} & Sampling
  offsets                                      \\
$\boldsymbol{Q}$ \textbf{}                 & Query tensor                                            \\
$\xrightarrow[]{}[d_1,\ d_2]$ \textbf{}    & Reshape to dimension~$d_1\times d_2$                    \\
\bottomrule
\end{tabular}
\end{table*}

\end{document}